\documentclass[journal]{IEEEtran}
\ifCLASSINFOpdf
\else
\fi

\usepackage{amsmath}
\usepackage{epsfig}
\usepackage{multirow}
\usepackage{color}
\usepackage[ruled,linesnumbered,noend]{algorithm2e}
\usepackage{algpseudocode} % or \usepackage{algcompatible}
\renewcommand{\algorithmiccomment}[1]{\bgroup\hfill\footnotesize$\triangleright$~#1\egroup}
\usepackage{placeins}

\usepackage{cite}
\usepackage{bm}
\usepackage{amsfonts}
\usepackage{graphics}
\usepackage{url}
\usepackage{soul}
\usepackage{mathrsfs}
\usepackage{flushend}
\usepackage{float}
\usepackage{tabularx}
\usepackage{array}
\usepackage{ragged2e}
\newcolumntype{P}[1]{>{\centering\hspace{0pt}}p{#1}} 
\newcolumntype{Z}{>{\centering\let\newline\\\arraybackslash\hspace{0pt}}X} 

\usepackage{lipsum} 

\soulregister{\cite}7
\soulregister{\citep}7
\soulregister{\citet}7
\soulregister{\ref}7
\soulregister{\pageref}7
\soulregister{\uppercase\expandafter}7
\sethlcolor{white}

\usepackage{threeparttable}

\begin{document}

%\title{DSP Map: A Dual-Structure Particle-based Occupancy Map for Dynamic Environments}
%\title{A Dual-Structure Particle-based Map for Continuous Occupancy Estimation in Dynamic Environments}
\title{Particle-based Instance-aware Semantic Occupancy Mapping in Dynamic Environments} %in Dynamic Environments 

\author{Gang~Chen\IEEEauthorrefmark{1}, Zhaoying~Wang\IEEEauthorrefmark{2}, Wei~Dong\IEEEauthorrefmark{2}, and Javier~Alonso-Mora\IEEEauthorrefmark{1} %, Wei~Dong, \text{X}injun~Sheng, ~\IEEEmembership{Member,~IEEE}, \text{X}iangyang Zhu, ~\IEEEmembership{Member,~IEEE} and Han Ding, ~\IEEEmembership{Member,~IEEE}% <-this % stops a space
\thanks{\IEEEauthorrefmark{1}The authors are with the Autonomous Multi-Robots Lab, Department of Cognitive Robotics, School of Mechanical Engineering,
        Delft University of Technology, 2628 CD, Delft, Netherlands. %E-mail: j.alonsomora@tudelft.nl.
        }
\thanks{\IEEEauthorrefmark{2}The authors are with the State Key Laboratory of Mechanical System and Vibration,
School of Mechanical Engineering, Shanghai Jiaotong University,
200240, Shanghai, China. %E-mail: j.alonsomora@tudelft.nl.
}
%\thanks{Corresponding author: Javier Alonso-Mora.
%        }
\thanks{This work is funded in part by the European Union (ERC, INTERACT, 101041863). Views and opinions expressed are however those of the author(s) only and do not necessarily reflect those of the European Union or the European Research Council Executive Agency. Neither the European Union nor the granting authority can be held responsible for them. 
        }
        }

%\author{
%    \IEEEauthorblockN{Author1\IEEEauthorrefmark{1}, Author2\IEEEauthorrefmark{2}, Author3\IEEEauthorrefmark{2}, Author4\IEEEauthorrefmark{1}}
%    \IEEEauthorblockA{\IEEEauthorrefmark{1}Institution1
%    \\\{1, 4\}@abc.com}
%    \IEEEauthorblockA{\IEEEauthorrefmark{2}Institution2
%    \\\{2, 3\}@def.com}
%}

%\markboth{Journal of \LaTe\text{X}\ Class Files,~Vol.~14, No.~8, August~2015}%

\maketitle

\begin{abstract}
Representing the 3D environment with instance-aware semantic and geometric information is crucial for interaction-aware robots in dynamic environments. 
Nevertheless, creating such a representation poses challenges due to sensor noise, instance segmentation and tracking errors, and the objects' dynamic motion.
This paper introduces a novel particle-based instance-aware semantic occupancy map to tackle these challenges. 
Particles with an augmented instance state are used to estimate the Probability Hypothesis Density (PHD) of the objects and implicitly model the environment. 
Utilizing a State-augmented Sequential Monte Carlo PHD (S$^2$MC-PHD) filter, these particles are updated to jointly estimate occupancy status, semantic, and instance IDs, mitigating noise. Additionally, a memory module is adopted to enhance the map's responsiveness to previously observed objects.
Experimental results on the Virtual KITTI 2 dataset demonstrate that the proposed approach surpasses state-of-the-art methods across multiple metrics under different noise conditions.
Subsequent tests using real-world data further validate the effectiveness of the proposed approach.
\end{abstract}
 

\IEEEpeerreviewmaketitle

\section{Introduction} \label{Section: Introduction} 
%\subsection{Motivation} \label{Section: Motivation}
Semantic mapping in unknown and unstructured environments \cite{kim20133d,yang2017semantic,sunderhauf_meaningful_2017,rosinol2021kimera,grinvald2019volumetric,naritapanopticfusion,seichter_panopticndt_2023} aims to represent both geometric and semantic information of elements utilizing onboard sensor data.
With the emergence of interaction-aware robots, e.g., robots that can interact with objects or other agents in the environment, it is essential to segment, track and model the individual instances with possible dynamic motions. The shape and motion of each instance should be updated consistently during the interactions to ensure the safety of the robot.  

While instance segmentation \cite{he2017mask,bolya2019yolact,lyu2022rtmdet} and tracking \cite{wen2021bundletrack,wang20206,wen2023bundlesdf} have been thoroughly explored in computer vision, the field of instance-aware semantic mapping in dynamic environments is still nascent.
This type of mapping poses considerable challenges: firstly, it demands the ability to manage noise not only from the sensor data but also from instance segmentation and tracking; secondly, it requires accounting for the dynamic motion of the instances, adding another layer of complexity to the task. 

Several methods have been proposed in recent years to realize instance-aware semantic mapping in static environments \cite{grinvald2019volumetric,naritapanopticfusion,seichter_panopticndt_2023}. However, the representations employed in these works, such as signed distance field or Gaussian kernels, do not account for real-time motion of objects in the environment, causing either missing object or trace noise problems \cite{MotionSC,dspMap}.
%In addition, the semantic or instance information is usually treated as a separate layer from the geometric information
%While our previous work \cite{dspMap} proposed a method to use particles to model the occupancy status of dynamic environments, it lacks the capability to estimate the semantics and instances of the objects. 
Alternatively, the occupancy status of dynamic environments can be modeled with particles \cite{SMCPHDMap,RFSMap,dspMap}. While effective in estimating the occupancy status, these methods do not incorporate the semantics and instances of the objects. 
Moreover, they consider each object to be composed of points with individual motions, thereby disregarding object-level motion and causing non-negligible noise in the occluded areas. 

In this work, we build upon the particle map work \cite{dspMap} by incorporating instance-aware semantic information and jointly updating the occupancy status, semantic, and instance labels using particles with an augmented instance state.
The proposed approach involves constructing a world model that represents different instances as distinct Random Finite Sets (RFSs) of points. 
The dynamic motions of each instance are addressed by sharing the same translation and rotation for the points within each set.
In accordance with the world model, an S$^2$MC-PHD filter is proposed to use particles to efficiently estimate the PHD of the RFSs while handling the aforementioned noise. The PHD works as an implicit representation, which is further used to estimate the occupancy status and semantic and instance IDs of each voxel subspace in the map.
\hl{In addition, we integrate a memory module into the filter to provide a conjecture of the occluded portion of a newly observed object, based on prior observations of objects with the same semantic label.}
%In addition, we introduce a memory module integrated into the filter to provide \hl{a conjecture of the occluded portion of a newly observed object, based on prior observations of objects with the same semantic label}, to enhance the map's responsiveness to the environment. 
In the experiments, the proposed method not only reaches state-of-the-art semantics and instance estimation performance in dynamic environment mapping but also improves the occupancy estimation performance by leveraging instance information.

%Tests using real-world data are also conducted to further validate the effectiveness of the proposed system.

The contributions of this paper are as follows:
\begin{itemize}
  \item An instance-aware point-based world model that enables using the PHD to implicitly represent the occupancy status as well as the semantic and instance IDs of the environment. %that represents the instances as point RFSs with 6D motions and
  \item An S$^2$MC-PHD filter that uses particles with an augmented instance state to efficiently estimate the PHD while handling the sensor and instance noise.
  \item Integrating an online memory module into filter-based mapping to enhance the map's responsiveness to previously observed objects.
  %A memory module integrated into the filter to enhance the map's responsiveness to previously observed objects.
  \item An efficient egocentric instance-aware semantic occupancy map that outperforms state-of-the-art maps in terms of occupancy, semantic, and instance estimation accuracy in the evaluated environments.
\end{itemize}

In terms of implementation, we also improved the data structure for particle-based mapping \cite{dspMap} to increase efficiency. Our code is available at \textit{\url{https://github.com/tud-amr/semantic_dsp_map}}. Video: \textit{\url{https://youtu.be/drSOYzVt2UM}}.

\section{Related Works} \label{Section: Related Work}

\subsection{Mapping in Dynamic Environments} \label{Section: Mapping in Dynamic Environments}
%Dynamic occupancy maps, including our previous DSP map. Compare the different representations. Nerf included.
Occupancy mapping in dynamic environments is a challenging task. %Traditional Bayesian update based \hl{voxel} map \cite{OctoMap,Ringbuffer,tordesillas2021faster} and signed distance field based map \cite{oleynikova2017voxblox,grinvald2019volumetric} usually suffer from the missing object or trace noise problem \cite{dspMap,MotionSC} due to the motions of the dynamic objects. 
\hl{Traditional mapping methods based on Bayesian updates \cite{OctoMap,Ringbuffer,tordesillas2021faster} and signed distance fields \cite{oleynikova2017voxblox,grinvald2019volumetric} usually suffer from the missing object or trace noise problem \cite{dspMap,MotionSC} due to the motions of the dynamic objects.}
The missing object problem occurs if a dynamic object is observed but not represented in the map and is mistakenly considered as free space. The trace noise problem occurs if a dynamic object has left an area while the map still considers (part of) the area as occupied. 

To address the problems, some works\hl{\cite{schmid_dynablox_2023,wu_moving_2024,schmid_khronos_2024,mersch2023building,8967704}} detect dynamic objects and eliminate them from the map. Then the dynamic objects are separately represented using the raw point cloud in the current frame, without considering modeling their motion or multi-view geometric information. 
To effectively model the motion and multi-view geometry of both dynamic and static objects in the map, enhanced representations of the environment are needed. One popular approach in the field of computer vision is to use neural radiance fields \cite{liu2022devrf,cao_hexplane_2023} or Gaussian splatting \cite{luiten2023dynamic,wu20234d} with additional time dimension to model each dynamic object.
Although these methods provide photo-realistic rendering of the environment, they require images from different views at different time steps in advance for training and are not suitable for real-time robotic tasks.
%These methods usually provide photo-realistic rendering of the environment but do not target to accurately model the geometric information. In addition, they require images from different views and time steps in advance for training and are not suitable for real-time robotic tasks. 

Another approach is to use particles as the representation \cite{SMCPHDMap,RFSMap,DynamicMapICRA2021,dspMap}. Compared to regular points in a point cloud, particles can be assigned a velocity vector and a weight, which can be used to model the motion and represent the uncertainty caused by noise. \hl{Danescu et al.} \cite{SMCPHDMap} first introduced the idea of using particles to model the objects and represent the occupancy status of the environment. \hl{Nuss et al.} \cite{RFSMap} improves the particle-based map by introducing RFS and using the PHD to represent the occupancy status. Further, our previous work \cite{dspMap} proposed to update the particles in the continuous space with a dual structure to improve the accuracy and efficiency of the map. These particle-based methods use particles with individual velocities to model the surface points on the objects. Particles representing the points on the same object can have very different velocities. While in the currently observed area, the particles with wrong velocities can be corrected by the measurements, the particles in the occluded area often cause noise in the map due to the ``particle false update'' problem \cite{dspMap}.
\hl{Compared to our previous work\cite{dspMap}, this paper introduces several key improvements. The most significant is the new S$^2$MC-PHD filter, which handles not only position noise but also object segmentation and tracking noise, whereas the previous filter\cite{dspMap} addressed only position noise. Another important aspect is that the S$^2$MC-PHD filter uses object-level transformations to model particle motions instead of individual particle velocities, solving the ``particle false update'' issue identified in\cite{dspMap} and reducing noise in occluded areas. Additionally, a memory module is introduced to enhance the map's responsiveness to previously observed objects, and the data structure is optimized to handle object-level information and increase mapping efficiency.
}

\subsection{Semantic Mapping} \label{Section: Semantic Mapping}
Semantic mapping represents the environment with both geometric and semantic information to realize better scene understanding and safer navigation. A number of works have studied semantic mapping in static environments \cite{kim20133d,yang2017semantic,sunderhauf_meaningful_2017,zeng_semantic_2018,yu_ds-slam_2018,nakajima2018efficient,grinvald2019volumetric,gan2020bayesian,rosinol2021kimera,seichter_efficient_2022,schmid_panoptic_2022,seichter_panopticndt_2023}. 
The representations used in these works include explicit representations, such as point cloud \cite{sunderhauf_meaningful_2017,zeng_semantic_2018}, voxel \cite{kim20133d,yang2017semantic,yu_ds-slam_2018}, mesh \cite{rosinol2021kimera}, etc. and implicit representations, such as signed distance field \cite{grinvald2019volumetric,schmid_panoptic_2022} and Gaussian kernels \cite{seichter_efficient_2022,seichter_panopticndt_2023}. 
To support the ability of planning interactions, several works \cite{grinvald2019volumetric,naritapanopticfusion,schmid_panoptic_2022,seichter_panopticndt_2023} added instance-aware information to the map as an additional layer. However, the motions of the instances are not considered and thus the maps are not suitable for dynamic environments. \hl{Recently, ConvBKI\cite{wilson2023convolutional,wilson2024convbki} combines the advantages of classical probabilistic algorithms and neural networks to build a semantic voxel map that can be used in dynamic environments, but the instance information is not considered.}
%many interaction-aware robotic tasks, the environment is dynamic, and the objects, such as vehicles, robots and humans, are moving. 
Realizing instance-aware semantic mapping in dynamic environments is still an open problem.

%Instance-aware semantic mapping in dynamic environments is valuable for interactions and is still an open problem.

%One approach \hl{related to} semantic mapping in dynamic environments involves employing occupancy networks \cite{li2023fb,ding2023multi,sima2023_occnet,tian_occ3d_2023,jiang2023symphonies}. 
%\hl{These methods address a related task but focus on predicting both visible and occluded areas in the current frame. }

\hl{Occupancy networks\cite{li2023fb,ding2023multi,sima2023_occnet,tian_occ3d_2023,jiang2023symphonies} are related to the task of semantic mapping in dynamic environments but focus on predicting both the visible and occluded areas in the current frame.
They currently lack the ability to retain memory of previously seen areas\cite{li2023sscbench} and are not instance-aware. 
Additionally, they face challenges in generalizing to new environments that differ from their training data.
In contrast, our approach belongs to the realm of classical mapping, where both visible areas from the current frame and memory of previously observed areas are considered. Particles are used to facilitate efficient instance-aware semantic mapping in dynamic environments, making it independent of specific scenes and capable of maintaining memory of previously observed areas.
}

\section{Preliminary} \label{Section: Preliminary}
Our map is built based on two concepts, RFS and SMC-PHD filter. This section briefly introduces these concepts as background knowledge.
Details can be found in \cite{PHD2003,2013ParticleFilterBook,RFSMap}. 

\subsection{Random Finite Set} \label{Section: Background RFS}
%\todo{Reorganize this part}

An RFS is a finite set-valued random variable \cite{RFSMap}. The number and the states of the elements within an RFS are random but finite. 
Let $\text{X}$ represent an RFS, and $\boldsymbol{x}^{(i)}$ denote the state vector of an element in $\text{X}$. Then $\text{X}$ is expressed as:
\begin{equation}\label{Eq: RFS intro}
  \text{X} = \left\{ \boldsymbol{x}^{(1)}, \boldsymbol{x}^{(2)}, ..., \boldsymbol{x}^{(N)} \right\}
\end{equation}
where $N \in \mathbb{N}$ is a random variable that represents the number of elements in $\text{X}$ and is referred to as the cardinality of $\text{X}$. Specifically, when $N=0$, $\text{X}$ is represented as the empty set, denoted by $\emptyset$. A typical application of RFS is in the field of multi-object tracking, where $\boldsymbol{x}^{(i)}$ typically represents the state of an object, and $\text{X}$ is a set composed of the states of all tracked objects. The value of $N$ varies as objects appear and disappear within the tracking range.

The first moment of an RFS is the PHD, which is used to describe the multi-object density. The PHD of $\text{X}$ at a state $\boldsymbol{x}$ is defined as follows: %The constant velocity model is utilized to predict the motion of the obstacles.
\begin{equation}\label{Eq: PHD}
  D_{\text{X}}(\boldsymbol{x}) = \mathbf{E} \left[ \sum_{\boldsymbol{x}^{(i)}\in \text{X}} \delta_d (\boldsymbol x-\boldsymbol{x}^{(i)}) \right] %= \int \sum_{\boldsymbol{x}^{(i)}\in \text{X}} \delta (x-\boldsymbol{x}^{(i)}) p_{\text{X}}(\text{X}) \delta \text{X}
\end{equation}
where $\delta_d(\cdot)$ is the Dirac delta function\footnote{Dirac delta function: $\delta_d(\boldsymbol{x})=0, \ \text{if} \ \boldsymbol x \neq \boldsymbol{0}$; $\int \delta_d(\boldsymbol{x}) \text d \boldsymbol{x}= 1$.} and $\mathbf{E}[\cdot]$ denotes the expectation.
The integral of the PHD corresponds to the expected cardinality of $\text{X}$, which can be expressed as:
\begin{equation}\label{Eq: PHD_integral}
  \int D_{\text{X}}(\boldsymbol{x}) \text{d}\boldsymbol x = \mathbf{E} [|\text{X}|]
\end{equation}
where $|\text{X}|$ is the cardinality of $\text{X}$. %This property is used to estimate occupancy status later.
Thus, higher PHD integral $\int D_{\text{X}}(\boldsymbol{x}) \text{d}\boldsymbol x$ suggests there are more elements in the RFS. If the $\boldsymbol{x}$ is constrained in a certain space in the map, it suggests that the space is more likely to be occupied by objects. Based on this property, the PHD can be used to estimate the occupancy status of the environment \cite{dspMap,RFSMap}.

% With the PHD working as an implicit representation, the occupancy status of a specified space can be estimated using the cardinality expectation \cite{dspMap} or the Pignistic probability \cite{RFSMap}.

% \cite{dspMap,RFSMap}. 

\subsection{SMC-PHD Filter} \label{Section: Background SMC-PHD Filter}
%\todo{Reorganize this part}
The PHD filter \cite{PHD2003} was introduced to predict and update the PHD of an RFS, \hl{and was originally designed for multi-object tracking. 
By updating the PHD rather than tracking and updating the state of each object, the PHD filter is more computationally efficient than Bayesian filters when the number of objects is large.}
The SMC-PHD filter is a PHD filter that utilizes the sequential Monte Carlo (SMC) method. Specifically, particles are used to represent the PHD. Each particle is usually characterized by a state vector that is composed of position and velocity, as well as an associated weight. Let $\text{X}_{k-1}$ represent the RFS at time $k-1$, $\tilde{\boldsymbol x}_{k-1}^{(i)}$ denote the state vector of the $i$-th particle, and $w_{k-1}^{(i)}$ denote the weight of the particle. Then the PHD of $\text{X}_{k-1}$ at state $\boldsymbol{x}_{k-1}$ can be approximated as:

\begin{equation}\label{Eq: PHD particles}
  D_{\text{X}_{k-1}}(\boldsymbol{x}_{k-1}) \approx \sum_{i=1}^{L_{k-1}} w_{k-1}^{(i)} \delta_d(\boldsymbol{x}_{k-1} - \tilde{\boldsymbol x}_{k-1}^{(i)})
\end{equation}
where $L_{k-1}$ is the number of particles at $k-1$. \hl{The approximation is utilized due to the statistical nature of the Monte Carlo method. Since achieving a good approximation requires a large number of particles, computational efficiency becomes essential for this filter.} 

The SMC-PHD filter estimates the PHD through prediction and updating of particles. The prediction step predicts the states changing from $k-1$ to $k$ and is described as:
\begin{equation}\label{Eq: prediction SMC-PHD}
    D_{\text{X}_{k|k-1}}(\boldsymbol{x}_k) = \sum_{i=1}^{L_{k-1}}P_s w_{k-1}^{(i)}  \pi_{k|k-1}(\boldsymbol{x}_k|\tilde{\boldsymbol x}_{k-1}^{(i)}) + \gamma_{k|k-1}(\boldsymbol{x}_k) 
\end{equation}
where $P_s$ is the survival probability \hl{indicating the probability that an object persists from $k-1$ to $k$}, and $\pi_{k|k-1}(\boldsymbol{x}_k|\tilde{\boldsymbol x}_{k-1}^{(i)})$ is the state transition density of the $i$-th particle. \hl{$\gamma_{k|k-1}(\boldsymbol{x}_k)$ is the intensity of the birth RFS. The birth RFS models the newly appeared objects in the tracking range at each time step, and its intensity controls the expected number of new objects.}

The update step updates the weight of each particle with measurements $\boldsymbol{z}_k \in \text{Z}_k$ and consequently updates the PHD with the following equations:
\begin{align}
  \label{Eq: particles_posterior_weights1}
  w_k^{(i)} = \left[ 1-P_d + \sum_{\boldsymbol{z}_k \in \text{Z}_k} \frac{P_d g_{k}(\boldsymbol{z}_k|\tilde{\boldsymbol x}_{k}^{(i)})}{ \kappa_k(\boldsymbol{z}_k) + \text{C}_k(\boldsymbol{z}_k) } \right] & w_{k|k-1}^{(i)} \\
  \label{Eq: particles_posterior_weights2}
  \text{C}_k(\boldsymbol{z}_k) = \sum_{j=1}^{L_k} P_d w_{k|k-1}^{(j)} g_{k}(\boldsymbol{z}_k|\tilde{\boldsymbol x}_{k}^{(j)}) & \\
  \label{Eq: particles_posterior_weights3}
  D_{\text{X}_{k}}(\boldsymbol{x}_k) \approx \sum_{i=1}^{L_{k}} w_{k}^{(i)} \delta(\boldsymbol{x}_k - \tilde{\boldsymbol x}_{k}^{(i)}) &
\end{align}
where $P_d$ is the detection probability \hl{that models the probability of an object being detected by the sensor},
$\text{Z}_k$ is the set of measurements at time $k$, $g_{k}(\boldsymbol{z}_k|\tilde{\boldsymbol x}_{k}^{(i)})$ is the likelihood and $\kappa_k(\boldsymbol{z}_k)$ is the clutter intensity, which represents the density of false measurements in the observation. %The clutter intensity is used to prevent the weight of the particles from being too small when there are no measurements. 
%The birth RFS is generated by the birth intensity $\gamma_{k|k-1}(\boldsymbol{x}_k)$.
Eq. (\ref{Eq: particles_posterior_weights1}) and Eq. (\ref{Eq: particles_posterior_weights2}) express the weight update of the particles. Eq. (\ref{Eq: particles_posterior_weights3}) describes the PHD of $\text{X}_k$ after the update step. \hl{The approximation is used for the same reason as in Eq. (\ref{Eq: PHD particles}).}
The number of particles $L_k$ usually differs from $L_{k-1}$ due to the birth and death of particles. %The birth and resampling module is used to generate new particles and prevent degeneracy. %Details can be found in \cite{PHD2003,2013ParticleFilterBook,RFSMap,dspMap}.

\section{World Model and System Structure} \label{Section: Map Overview}
\subsection{World Model} \label{Section: World Model}

\begin{figure*}
  \centering
  \includegraphics[width=7.2in]{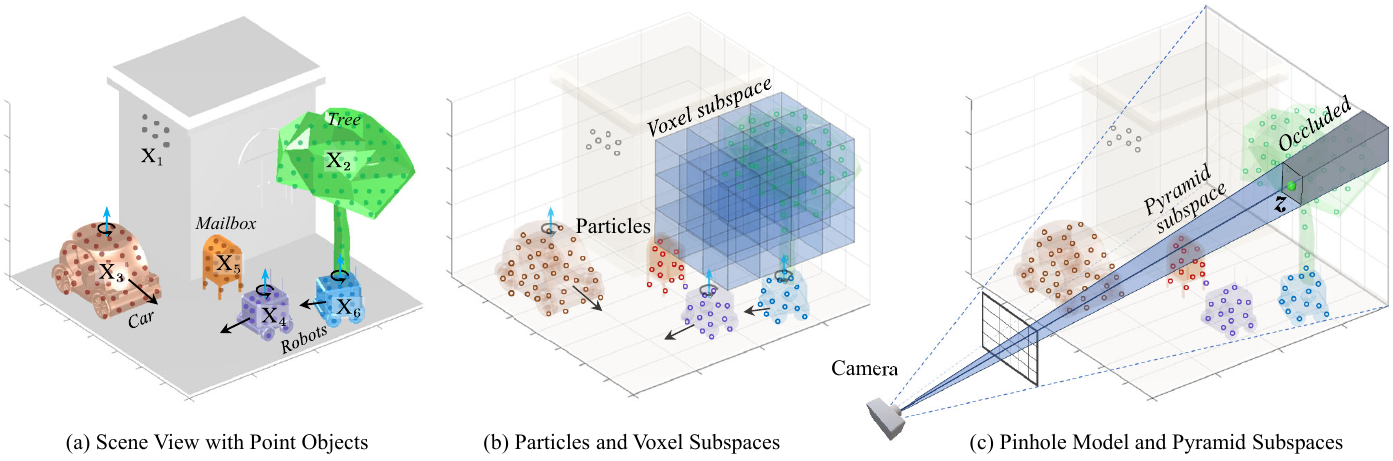}
  \caption{World model illustration. (a) shows an example scenario with both objects of interests and background objects. The wall and ground belong to the background object with ID 1. The tree belongs to the background object with ID 2. The car and the two robots belong to the objects of interest with ID 3, 4 and 6 respectively, and are dynamic objects. The mailbox with ID 5 is an example static object of interests. Each object is assumed to be composed of a set of points on its surface. The set of the $I$-th object is marked as $\text{X}_I$ and the points within the set share the same color. (b) shows the particles (hollow points) that are used to model the PHD of the points. Particles with different IDs are shown in different colors. Particles with the same ID share the same motion. The particles are stored in voxel subspaces \cite{dspMap}, which are also used for resampling and occupancy estimation. (c) shows the camera Pinhole model used in this work to formulate the pyramid subspaces \cite{dspMap}, which are used to distinguish the observed area and occluded area in the continuous space and to accelerate the update process. The green point is a measurement point in a pyramid subspace. The gray area behind the measurement point is occluded. Only a part of the points in $\text{X}_1$ in (a), voxel subspaces in (b), and pyramid subspaces in (c) are shown for clear illustration. %\todo{Set ticks. Make (a) looks better. Missing vertical line in (c).}
  }
  \label{Fig: world model}
\end{figure*}

We assume the map space contains only rigid objects, which may be moving. 
Each object has an instance ID and has a random but finite number of points representing its shape. These points are on the surface of the object and are usually observed as point cloud. Fig. \ref{Fig: world model} (a) illustrates a scene with objects and points on each object. By assuming the rigidity, all the points of an object share the same motion between two time steps. 
We do not specifically model non-rigid objects, such as pedestrians, but treat them as rigid objects with a simplified motion model. 

%The shape changing caused by limb motion is tackled by  
%as a pragmatic approximation.
%For non-rigid objects, such as pedestrians, the assumption of rigid body motion is also employed and the shape changing caused by limb motion is tackled by . 
%as a pragmatic approximation 

To associate each point with its corresponding object and estimate the semantics and instances, we use a state vector composed of the 3D position and an augmented instance ID dimension for each point. The state vector is expressed as:
\begin{equation} \label{Eq: point state vector}
 \boldsymbol{x} = \left[ x, y, z, id \right]^T
\end{equation}
where $id \in \mathbb{N}^{+}$ is the instance ID, and $x$, $y$, $z \in \mathbb R$ are the 3D position coordinates of the point in the map space, which is a cubic space \hl{centered at the location of the robot}. While $id$ is a discrete state variable, it can be regarded as a narrowed continuous state variable, and the Dirac delta function $\delta_d(\cdot)$ to calculate the PHD can still be used. 
%Each instance ID corresponds to a semantic class when the instance is created.
\hl{Each instance is assigned with a semantic label when the instance is created.}
%The rest functions that contain $\boldsymbol{x}$ are presented in Section \ref{Section: S2MC-PHD Filter}.

All the points in the map form an RFS, which can be represented as:
\begin{equation}
  \label{Eq: X sub-RFS}
  \begin{aligned}
  \text{X}= & \{\underbrace{\boldsymbol{x}^{(1)}, \boldsymbol{x}^{(2)}, \cdots, \boldsymbol{x}^{\left(n_1\right)}}_{\text{X}_1}, \underbrace{\boldsymbol{x}^{\left(n_1+1\right)}, \cdots,
  \boldsymbol{x}^{\left(n_1+n_2\right)}}_{\text{X}_2}, \cdots, \\ & \underbrace{\boldsymbol{x}^{\left(n_1+n_2+\cdots, n_{N-1}+1\right)}, \cdots, \boldsymbol{x}^{\left(n_1+n_2+\cdots, n_{N}\right)}}_{{\text{X}}_N} \}
  \end{aligned}
\end{equation}
%\todo{Remove the brace.}
%where $\text{X}_I, I \in \{1, 2, \cdots, N\}$ under the braces is a sub-RFS composed of the points belonging to object $I$, $n_I$ is the number of points in $\text{X}_I$, and $N$ is the number of objects in the map. $n_I$ changes at each time step as new parts of the $I$-th object are observed or some parts of the object are out of the map range. $N$ also changes as objects enters and moves out of the map range. For points in $\text{X}_I$, their $id$ is always $I$. Fig. \ref{Fig: world model} (a) illustrates the objects, points and the corresponding $\text{X}_I$ in a scenario.
where $\text{X}_I, I \in \{1, 2, \cdots, N\}$ under each brace is a sub-RFS composed of the points belonging to instance $I$. $n_I$ represents the number of points in $\text{X}_I$, and $N$ denotes the total number of instances present in the map. The value of $n_I$ changes at each time step, reflecting the observation of new part of the $I$-th instance or the removal of part of the instance from the map. 
Similarly, the value of $N$ changes as objects enter or exit the map's boundaries. For the points within $\text{X}_I$, their associated $id$ is always $I$. %Figure \ref{Fig: world model} (a) provides an illustration of objects, points, and their corresponding $\text{X}_I$ in a given scenario.

The instances can be categorized into two groups: instances of interest and background instances. Instances of interest may exhibit either static or dynamic behavior at one moment. Each instance of interest has a distinct ID and is segmented and tracked. 
%The points associated with an instance of interest share identical motions, which is, from time $k-1$ to $k$, the coordinates of the points in $\text{X}_I$ are transformed with the same transformation matrix $\boldsymbol{T}_{I,k \ 4\times 4}$. 
%The points in $\text{X}_I$ share the same transformation matrix.
%identical linear and angular velocities at any given time, which is
%For an object with the ID $I$, the velocities of points in $\text{X}_I$ are as follows:
%\begin{equation} \label{Eq: object velocity}
%$ \boldsymbol{V}_I = \left[ \boldsymbol{v}_I, \boldsymbol{\omega}_I \right]^T$, 
%\end{equation}
%where $\boldsymbol{v}_I$ is the linear velocity and $\boldsymbol{\omega}_I$ is the angular velocity. \todo{ADD T}
%When the instance is a non-rigid object, such as a pedestrian, we still assume that all the associated points share identical velocities as we don't consider sub-instance semantics. 
%\todo{Consider if change the phrase.}
%We assume that each instance can be detected and segmented. Their IDs are assigned dynamically by an object-level tracker.
The background instances, encompassing both unlabeled objects and labeled objects that are not of interest, are static. In the context of a navigation task, background instances may refer to features like the ground, walls, trees, etc. 
%unlabeled objects may refer to features like the ground, walls, and other unclassified elements, while labeled static objects that are not of interest could include trees, cabinets, etc.
We assume the same kind of background objects have the same instance ID and belong to one sub-RFS. For example, we can presume that all walls in the map belong to $\text{X}_1$, and all trees in the map belong to $\text{X}_2$.

%Our mapping approach involves updating the PHD of $\text{X}$ to model the objects within the map space. 
The measurements of $\text{X}$ consist of point clouds accompanied by their respective instance IDs. These points are typically acquired from a stereo/RGB-D camera or Lidar. The instance IDs are derived from instance segmentation and tracking. The measurements have the similar form as $\text{X}$:
\begin{equation}
  \begin{aligned}
  \text{Z}= & \{\underbrace{\boldsymbol{z}^{(1)}, \boldsymbol{z}^{(2)}, \cdots, \boldsymbol{z}^{\left(m_1\right)}}_{\text{Z}_1}, \underbrace{\boldsymbol{z}^{\left(m_1+1\right)}, \cdots,
  \boldsymbol{z}^{\left(m_1+m_2\right)}}_{\text{Z}_2}, \cdots, \\ & \underbrace{\boldsymbol{z}^{\left(m_1+m_2+\cdots, m_{M-1}+1\right)}, \cdots, \boldsymbol{z}^{\left(m_1+m_2+\cdots, m_{M}\right)}}_{{\text{Z}}_M}\}
  \end{aligned}
\end{equation}
where $\text{Z}_J, J \in \{1, 2, \cdots, M\}$ under each brace is a sub-RFS composed of the measurement points belonging to object $J$, $m_J$ is the number of points in $\text{Z}_J$, and $M$ is the number of objects observed in the measurements of this time step. Each point in the measurement is also composed of a 3D position and an instance ID:
\begin{equation} \label{Eq: point measurement}
  \boldsymbol{z} = \left[ x, y, z, id \right]^T
\end{equation}

%To estimate the object-level velocity, we assume that from time $k-1$ to $k$, the transition and orientation of an object in tracking can be estimated and represented as a transformation matrix. For example, $\boldsymbol{T}_{J,k \ 4\times 4}$ represents the transformation matrix of object $J$ from time $k-1$ to $k$. 

Note that due to the sensor's limited field of view (FOV) and inevitable occlusions between objects. Only a portion of objects and points in $\text{X}$ has observations in $\text{Z}$. 
Furthermore, the measurements contain noise in both position and instance ID. The noise in position comes from the sensor measurement noise. 
The noise in instance ID arises from missing instances, merging instances, misclassification, etc., in instance segmentation, and incorrect data association in tracking, and manifests as the incorrect instance IDs. %: 1) incorrect IDs of points in a whole subset $\text{Z}_I$; 2) some points belonging to object $I$ being erroneously grouped with object $J$, where $I\neq J$. 
When updating the map, noise in position and instance should both be addressed.

%We can further categorize the noise into object-level noise and point-level noise. The position noise and 2) are point-level noise. 1) is object-level noise.
%When updating the map, both object-level noise and point-level noise should be considered.

Our map employs particles with IDs to approximate the Probability Hypothesis Density (PHD) of $\text{X}$, and $\text{Z}$ is utilized to update the particles. \hl{Compared to using particles to track and update the state of each point individually, modeling the PHD is more computationally efficient and can better handle occlusions.} 
We illustrate the particles in Fig. \ref{Fig: world model} (b) and (c) with hollow points. Following our previous work \cite{dspMap}, we store the particles in voxel subspaces and use pyramid subspaces to differentiate between the currently observed and occluded areas in the continuous space. %The particles in the occluded area are not updated.
%The particles in the currently observed area are updated with the measurements, while the particles in the occluded area are not updated. %The particles in the occluded area are not updated.
%To enhance the handling of image input and expedite the update process, 
To realize efficient pyramid subspace division when the robot is moving, this paper adopts the Pinhole model for pyramid subspace division, as illustrated in Fig. \ref{Fig: world model} (c). Details about the subspace divisions can be found in Section \ref{Section: Data Structure}. The following presents the system structure and the filter used to update the PHD in the observed area of each time step.

%The particles within a pyramid subspace can be projected onto a pixel on the image plane. Each pixel corresponds to a measurement point, with the space behind the measurement point being occluded, while the space in front of the measurement point is visible. The particles in the occluded space are not updated.

\subsection{System Structure} \label{Section: System Structure}

\begin{figure}[t]
  \centering
  \includegraphics[width=3.5in]{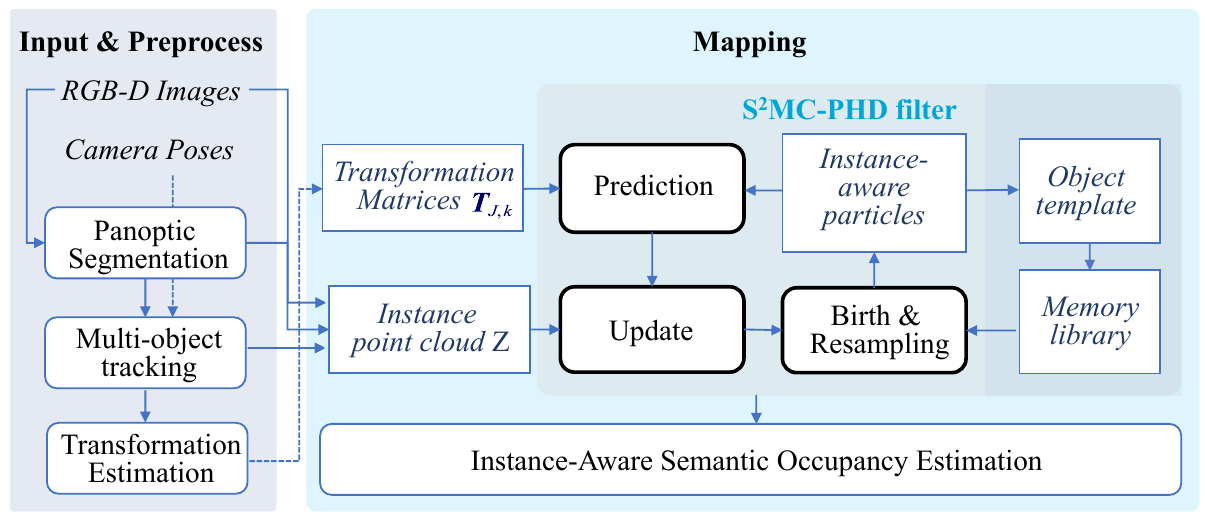}
  \caption{System structure. The left side shows the input and preprocessing modules, which generate data composed of two parts: transformation matrices and instance point cloud. The generated data, which contains noise, is used in the mapping on the right side. The core of the mapping part is the S$^2$MC-PHD filter.}
  \label{Fig: system structure}
\end{figure}

The system structure is shown in Fig. \ref{Fig: system structure}. The modules on the left detail the input and preprocessing requirements.
The inputs are RGB-D image pairs from stereo / RGB-D cameras\footnote{Using Lidar point cloud can also acquire the input required in our world model. As panoptic segmentation and tracking are more accessible with RGB images, we take RGB-D images as input in this paper.} and the corresponding poses of the camera.
The preprocessing modules encompass panoptic segmentation, multi-object tracking and transformation estimation.
At each time step, the panoptic segmentation module \cite{he2017mask,bolya2019yolact,lyu2022rtmdet} segments the instances in the image, and the multi-object tracking module \cite{maggiolino2023deep,xu2022pp} tracks the instances of interest in the image sequence.
\hl{Recent advancements in 4D panoptic segmentation have proposed an alternative solution for obtaining both segmentation and tracking results using a single network \cite{aygun20214d}.}
The transformation estimation module estimates the motion of instances in tracking between two frames and can be realized by joint localization and object motion estimation \cite{huang2020clustervo,bescos2021dynaslam,qiu2022airdos} or pose tracking methods \cite{wen2021bundletrack,wang20206,wen2023bundlesdf}.
We consider these preprocessing modules off-the-shelf and do not delve into them in this paper. However, these modules inevitably introduce noise to the measurements, which should be considered when updating the map. 
In Section \ref{Section: Implementation Details}, we present a practical implementation solution for the modules. The output of the preprocessing modules encompasses the measurement point RFS $\text{Z}$, and the estimated transformation matrices of objects of interest between two consecutive frames.

% which are used later in the subsequent prediction step.

% The measurement set $\text{Z}$ can be generated by making point cloud with the depth image and localization result \cite{geneva2020openvins,qin2018vins}, and giving IDs to the point cloud with instance segmentation \cite{lyu2022rtmdet,woo2023convnext} and multi-object tracking \cite{maggiolino2023deep,xu2022pp}. 
% The estimation of the instance movement can be realized by localization methods that consider joint object motion estimation \cite{huang2020clustervo,bescos2021dynaslam,qiu2022airdos} or by additional pose tracking methods \cite{wen2021bundletrack,wang20206,wen2023bundlesdf}.
%  %and use them to realize mapping with a real robot.

%For example, the localization can be realized with visual inertial odometry \cite{geneva2020openvins,qin2018vins}, and the instance segmentation and multi-object tracking can be realized with recent advances \cite{lyu2022rtmdet,woo2023convnext,maggiolino2023deep,xu2022pp} in the field of computer vision. In the experiment section, we present a solution of the preprocess modules and use them to realize mapping with a real robot.

%The mapping part comprises two main components: the S$^2$MC-PHD filter and the memory module. 
An S$^2$MC-PHD filter, detailed in the next section, is proposed to \hl{update the PHD of $\text{X}$ at the sub-instance level, and estimate the occupancy status of the map.}
%The S$^2$MC-PHD filter is designed to employ instance-aware particles to efficiently approximate the $\text{X}$'s PHD, which is then used to calculate the occupancy status of the map. 
This filter mainly consists of particle prediction, update and birth and resampling modules. The prediction module adopts the transformation matrices of the objects of interest to predict the new positions of the points. 
The update module uses the measurements $\text{Z}$ to update the PHD represented by particles. With the updated PHD, the occupancy status and instance labels of the map can be estimated.
The birth and resampling module generates new particles and prevents degeneracy.
The memory module is introduced to enhance the map's responsiveness to previously observed objects and to provide a conjecture of the occupancy status in the occluded portion of an object.

\section{S$^2$MC-PHD Filter} \label{Section: S2MC-PHD Filter}
The S$^2$MC-PHD Filter is built upon the world model described in Section \ref{Section: World Model} and uses particles to approximate the PHD of $\text{X}$. 
In addition to the position state and weight, we augment the state vector of a particle in the SMC-PHD filter with an instance ID. Each particle can be regarded as a hypothesis of the point in the world model. A particle with index $i$ at time step $k$ is represented by:
\begin{equation} \label{Eq: particle state vector}
  P^{(i)}_k = \left\{ \tilde{\boldsymbol x}^{(i)}_k, w^{(i)}_k \right\} = \left\{  \left[x^{(i)}_k, y^{(i)}_k, z^{(i)}_k, id^{(i)} \right]^T, w^{(i)}_k \right\}
\end{equation}
In Fig. \ref{Fig: world model} (b) and (c), the particles are shown with hollow points. Different colors indicate different instance IDs.
%We therefore assume that one particle can only represent the PHD of one sub-RFS composed of the points with the same instance ID in Eq. (\ref{Eq: X sub-RFS}).
% The subscript $k$ is not added to $id^{(i)}$ because it is a discrete 
%the instance ID of a particle is created when the particle is born and is not changed afterwards. 

% where $id^{(i)}$ is the instance ID of the particle. 
% We assume that one particle can only represent the PHD of the instance with the same ID. 

%The PHD of $\text{X}$ at time $k$ is approximated by:
% \begin{equation}
%   D_{\text{X}_k}(\boldsymbol{x}_k) \approx \sum_{i=1}^{L_k} w^{(i)}_k \cdot \delta_d ( \boldsymbol{x}_k - \tilde{\boldsymbol x}^{(i)}_k  )
% \end{equation}
% where 
%$\delta_{kr}(\cdot)$ is the Kronecker delta function\footnote{Kronecker delta function: $\delta_{kr}(x, y)=1, \ \text{if} \ x=y$; $\delta_{kr}(x, y)=0, \ \text{if} \ x \neq y$.}, which is used to omit the particles that belong to other instances. 
% $L_k$ is the number of particles at time $k$.

\subsection{Prediction} \label{Section: Prediction}
The prediction step predicts the PHD distribution of $\text{X}$ based on the estimated instances' transformation matrices given by the preprocessing module. By using the transformation matrices, the 6D motion of the instance between two time steps can be tackled.
Let $\boldsymbol{T}_{J,k}$ denote the transformation matrix estimated for the $J$-th observed instance at time $k$. The transformation matrix contains the rotation matrix and translation vector and is a $4\times4$ matrix. 
%The transformation matrix is estimated by the transformation estimation module in Section \ref{Section: World Model}. The transformation matrix is used to predict the motion of the points in the instance. %The transformation matrix is used to predict the motion of the points in the instance.
%\todo{Add the statement of the transformation matrix here.}
% Since we have assumed that the objects in the map space are rigid, the points of one object should follow the same transformation matrix. Let $^{I}{\mkern\supscriptkern}\boldsymbol{T}_k$ denote the transformation matrix estimated for object $I$ at time $k$. The transformation matrix contains the rotation matrix and translation vector and is a $4\times4$ matrix. 
Let $f(\boldsymbol{T}_{J,k}, \boldsymbol{x})$ denote the function that uses $\boldsymbol{T}_{J,k}$ to transform a point $\boldsymbol{x}$ that belongs to object $J$ from time $k-1$ to $k$.
Then the predicted prior state of this point is:
\begin{equation} \label{Eq: prediction object}
  \begin{aligned}
    \boldsymbol{x}_{k|k-1} & = f(\boldsymbol{T}_{J,k}, \boldsymbol{x}_{k-1}) + [\boldsymbol{\xi}, 0]^T \\
    & = \boldsymbol{T}_{J,k} \begin{bmatrix} \boldsymbol{x}_{k-1}(1:3) \\ 1 \end{bmatrix} + \begin{bmatrix} \boldsymbol{0}_{3\times 1} \\ \boldsymbol{x}_{k-1}(4) \end{bmatrix} + \begin{bmatrix} \boldsymbol{\xi} \\ -1 \end{bmatrix}
  \end{aligned}
\end{equation} 
where $\boldsymbol{\xi}$ represents the position noise caused by the inaccuracy in the transformation matrix estimation.
We assume the noise follows a Gaussian distribution with a zero mean and a covariance matrix $\boldsymbol{Q}_{3 \times 3}$, denoted as $\boldsymbol{\xi} \sim \mathcal{N}(\boldsymbol{0}, \boldsymbol{Q})$.

Then, the state transition function from a particle at time $k-1$ to a prior point state at time $k$ can be formulated as a Gaussian probability density function:
\begin{equation}
  \pi_{k|k-1}(\boldsymbol{x}_k|\tilde{\boldsymbol x}_{k-1}^{(i)}) = \mathcal{N}\left(\boldsymbol{x}_k; f(\boldsymbol{T}_{J,k}, \tilde{\boldsymbol x}_{k-1}^{(i)}), \boldsymbol{Q}\right)
\end{equation}
By substituting the state transition function into Eq. (\ref{Eq: prediction SMC-PHD}), the predicted PHD of $\text{X}$ can be obtained.

% At the particle level, .

Due to the limited FOV of the sensor and inevitable occlusion between objects, some objects cannot be observed, and their transformation matrices are not available. In this case, we employ a constant velocity model to predict the transformation matrix. \hl{The ego motion of the sensor is handled with the data structure described in Section \ref{Section: Data Structure}}.
% If an object is visually re-tracked after being occluded for a while, the positions predicted by the constant velocity model may be very different from the actual positions due to the complex motion of real objects. 
% While the update step and the particle birth step can correct this prediction error naturally by lowering the weight of the predicted particles in the wrong positions and generating new particles at the correct positions, the observed information in the previous tracking of the object is lost. To address this issue, we utilize the memory module to transform the instance's particles and align them with the re-tracked measurements in order to leverage the information in previous tracking. Details can be found in Section \ref{Section: Memory Enhancement}.

\subsection{Update} \label{Section: Update}
The update procedure updates the PHD distribution of $\text{X}_k$ by calculating the particle weights using the latest measurements $\text{Z}_k$ at time $k$. The update is performed only for the particles in the visible space.
In this subsection, our primary focus is on updating the particle weight utilizing measurements $\text{Z}_k$ while mitigating the effects of noise discussed in Section \ref{Section: World Model}. %Additionally, we explore the efficiency of the update process by incorporating Pinhole model-based pyramid subspaces.

The miss-detection and clutter noise in raw sensor measurements has been taken into account by the detection probability $P_d$ and the clutter intensity $\kappa_k(\boldsymbol{z}_k)$ in the original SMC-PHD filter introduced in Section \ref{Section: Background SMC-PHD Filter}. If the panoptic segmentation and tracking are very reliable, and consequently, there is no instance noise, the weight of the particle with the augmented state vector can be updated using a straightforward method, i.e., \textbf{Individual Filtering (IF)} method: updating the particles exclusively with measurements sharing the same instance ID. In other words,
particles with $id^{(i)} = J$ are updated only with the measurements in $\text{Z}_J$.
Essentially, the filter comprises multiple independent SMC-PHD filters, where each filter works for an RFS of a specific instance in Eq. (\ref{Eq: X sub-RFS}). 

\begin{figure}[t]
  \centering
  \includegraphics[width=3.2in]{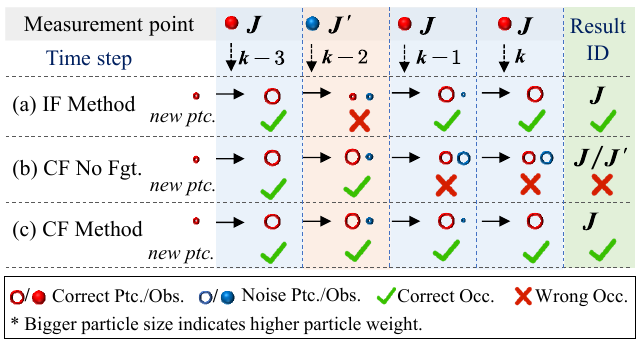}
  \caption{Illustration of the updated particles of filters (a) to (c) when a noise observation with a wrong ID is given.  
  With Method IF, the particles' weights at $k-2$ are too small, and the occupancy status of the space is thus falsely treated as free. With Method CF No Forgetting function, the instance ID after $k-1$ is ambiguous.  %The particles are shown with hollow points, where a larger size indicates a higher weight. Initially, one particle with ID $J$ (red) with a small newborn weight exists. 
  %After a measurement point is given, each particle's weight is updated. The measurement point at $k-2$ is mislabeled as $J^\prime$ (blue). A new blue particle with ID $J^\prime$ is created during the update. 
  %The tick and cross under each result after the update indicate whether the occupancy status of the point is correctly estimated.
  }
  \label{Fig: particle_id_update}
\end{figure}

However, if there is instance noise, the IF method suffers from a missing object problem.
For example, when some measurement points $\text{Z}_{J}$ of instance $J$ are mislabeled with another existing or new instance ID $J^\prime$ due to the misclassification, inaccurate segmentation or wrong data association, the weights of the particles whose $id^{(i)} = J$ will be decreased. 
Simultaneously, the particles whose $id^{(i)} = J^\prime$ will be created in the particle birth step (Section \ref{Section: Birth and Resampling}) but will only have a low weight. Consequently, there is a sudden drop in the PHD at the positions of these points. The region is susceptible to being inaccurately classified as free space, posing a high risk of collision for the robot. %Particularly, in cases where mislabeling occurs frequently, the resulting map becomes highly unreliable.
We illustrate this issue in Row (a) of Fig. \ref{Fig: particle_id_update} with a single measurement point and single particle situation. At $k-2$, the measurement point on an instance is mislabeled, and the estimated occupancy result at this step is wrong because both particles have very low weights. The occupancy status of the space where the point is located will be falsely estimated as free space.

To address the instance noise, we further propose \textbf{Collective Filtering (CF)} method: updating the particles collectively with all the measurements $\text{Z}_k$ by using a specialized likelihood function. %The likelihood function is designed to handle both the position and the instance noise.
%and the augmented state vector with instance ID. %The solution is to design a likelihood function that can tackle both the position and the instance noise.
The likelihood function is formulated as:
\begin{equation}\label{Eq: likelihood new}
  g_{k}(\boldsymbol{z}_k|\tilde{\boldsymbol x}_{k}^{(i)}) = F_{gt}(\tilde{\boldsymbol x}_{k}^{(i)})  \cdot T_r(\boldsymbol{z}_k, \tilde{\boldsymbol x}_{k}^{(i)}) \cdot \mathcal{N}\left(\boldsymbol{z}_k; \tilde{\boldsymbol x}_{k}^{(i)}, \boldsymbol{\Sigma}\right)
\end{equation}
where $T_r(\cdot)$ represents an instance ID transition function and $F_{gt}(\cdot)$ is a forgetting function. $\mathcal{N}(\cdot)$ is the Gaussian probability density for position transition, used to model the position noise, whose covariance matrix is assumed to be $\boldsymbol{\Sigma}$.

$T_r(\boldsymbol{z}_k, \tilde{\boldsymbol x}_{k}^{(i)})$ is defined as:
\begin{equation} \label{Eq: instance ID transition function}
  T_r(\boldsymbol{z}_k, \tilde{\boldsymbol x}_{k}^{(i)}) = \begin{cases}
    1, & \text{if} \ \tilde{\boldsymbol x}_{k}^{(i)}(4) = \boldsymbol{z}_k(4) \\
    P_{tr}\left(\boldsymbol{z}_k(4), \tilde{\boldsymbol x}_{k}^{(i)}(4)\right), & \text{Otherwise}
  \end{cases}
\end{equation}
where $P_{tr}\left(\boldsymbol{z}_k(4), \tilde{\boldsymbol x}_{k}^{(i)}(4)\right) \in [0,1)$ characterizes the likelihood of an instance being identified as or associated with another instance. This parameter can be determined as a function of instance labels and positions if the performance of instance segmentation and tracking is known. For generality and computational simplicity, we treat it as a constant in the experiments.

The forgetting function $F_{gt}(\tilde{\boldsymbol x}_{k}^{(i)})$ is elaborated as a truncated Ebbinghaus Curve of Forgetting \cite{ebbinghaus2013memory}:
\begin{equation} \label{Eq: forgetting function}
  F_{gt}(\tilde{\boldsymbol x}_{k}^{(i)}) = \begin{cases}
    e^{- \frac{\Delta k^{(i)}}{S}}, & \text{if} \ \Delta k^{(i)} \leq \Delta \bar{k} \\
    0, & \text{if} \ \Delta k^{(i)} > \Delta \bar{k}
  \end{cases}
\end{equation}
where $e$ represents the Euler's number. $\Delta k^{(i)} \in \mathbb{N}$ denotes the time interval between the current time step $k$ and the last time step when the $i$-th particle was updated with a measurement sharing the same ID. $\Delta \bar{k}$ is a threshold that controls the maximum time interval that the particle can be updated with the measurement that has a different ID. The constant $S > 0$ governs the forgetting speed, with a smaller $S$ resulting in a faster rate of forgetting.

% {\color{blue}We assume that the position error is independent on each axis. Then the position transition function can be formulated as a product of three independent Gaussian PDFs:}
% \begin{equation}
%   \mathcal{N}\left(\boldsymbol{z}_k; \tilde{\boldsymbol x}_{k}^{(i)}, \boldsymbol{\Sigma}\right) = \prod_{i \in \{{x},{y},{z} \}^{}} \mathcal{N}\left(z_{k,i}; p_{k,i}, \rho^2 (r_k) \right)
% \end{equation}

The ID transition function allows particles with a different ID from the measurement to still be updated if the measurement's position is close. Then, the aforementioned missing object problem can be avoided. 
However, if an object with ID $J$ is persistently labeled with ID $J^\prime$ afterward in the tracker or if it is relabeled as $J$ after mislabeled as $J^\prime$, both particles with ID $J$ and $J^\prime$ will have large weights, which causes confusion in labeling the space occupied by the object. 
We illustrate the situation where the object is mislabeled with ID $J^\prime$ and then relabeled as $J$ in Row (b) of Fig. \ref{Fig: particle_id_update}. The result contains ambiguous ID choices.
Therefore, the forgetting function becomes crucial. This function reduces the weight of particles not updated with measurements sharing the same ID. When $\Delta k^{(i)} > \Delta \bar{k}$, the particle's weight experiences a rapid and substantial decrease, finally being removed in the resampling step. As a result, the map turns to trust more on consistent measurements. If $J$ is permanently mislabeled as $J^\prime$, the particles with ID $J$ will be removed after a few observations, and the space taken by the object will only be labeled with $J^\prime$. If $P_{tr}$ is set to be zero and the forgetting function is not used, the CF method will be equivalent to the IF method.

%The speed of how fast the particles with ID $J$ are removed is controlled by $S$.

Updating each particle with each measurement is computationally expensive, as discussed in \cite{dspMap}. To accelerate the update process while considering the occluded space, we incorporate the Pinhole model-based pyramid subspaces and activation bounding boxes to confine the particles that should be updated with each measurement point. The details can be found in Section \ref{Section: Implementation Details} and the Appendix.

% The Gaussian position transition function in Eq. (\ref{Eq: likelihood new}) is assumed to exhibit independent noise on each axis, with the variance being identical for each axis: $\boldsymbol{\Sigma} = \rho^2 \boldsymbol{I}_{3 \times 3}$, where $\rho^2$ is the variance of each axis. 
% For a real depth camera, $\rho$ typically follows a quadratic relationship with the distance from $\boldsymbol{z}_k$ to the camera \cite{mallick2014characterizations,sweeney2019supervised}. Despite this quadratic relationship, given a $\boldsymbol{z}_k$ with a corresponding $\rho$, the particles that satisfy $\{ \tilde{\boldsymbol x}_{k}^{(i)} | \mathcal{N}(\boldsymbol{z}_k; \tilde{\boldsymbol x}_{k}^{(i)}, \boldsymbol{\Sigma}) \geq \epsilon \}$ are confined in a sphere in the 3D space. Here, $\epsilon$ is a threshold that is close to zero. In the Appendix, we derive the radius of this sphere and the formulation of the projected ellipse of the sphere onto the camera image plane. Utilizing this formulation, we further derive a bounding box for the ellipse. The bounding box plays the role of the activation space in \cite{dspMap}. Particles outside the bounding box are not updated to accelerate the update process. 
%Details of how to realize the update efficiently can be found in Section \ref{Section: Implementation Details}.

\subsection{Particle Birth, Resampling and Occupancy Estimation} \label{Section: Birth and Resampling}
Particles are typically born from the measurements $\text{Z}_k$. For each measurement point $\boldsymbol{z}_k \in \text{Z}_k$, we generate $L_b$ newborn particles. The state of each newborn particle $P_{b,k} = \left\{ \tilde{\boldsymbol x}_{b,k}, w_{b,k} \right\} $, where the subscript $b$ suggests ``born'', is given by the following equations:
\begin{align}
  \label{Eq: particle_birth_state1}
  & \tilde{\boldsymbol x}_{b,k} (1:3) = \boldsymbol{z}_k (1:3) + \boldsymbol{\sigma}, \boldsymbol{\sigma} \in \mathcal{N} (0, \boldsymbol{\Sigma})  \\
  \label{Eq: particle_birth_state2}
  & \tilde{\boldsymbol x}_{b,k} (4) = \boldsymbol{z}_k (4), \quad  w_{b,k} = \frac{v_{b,k|k-1}}{M_k L_b}  
\end{align}
where $v_{b,k|k-1} = \int \gamma_{k|k-1}(\boldsymbol{x}_k) \text{d} \boldsymbol{x}_k$ is a parameter that controls the expected number of newborn points from $k-1$ to $k$. 
The instance ID of a newborn particle is the same as the measurement point. Suppose the measurement point of an object is labeled differently at a new time, in which case instance noise occurs, the instance ID of the newborn particle will also change, and new instance hypotheses will be generated. These hypotheses are updated with the measurements in the subsequent time steps to filter out the incorrect ones with the ID transition function and the forgetting function. 
With the newborn particles, Equations (\ref{Eq: particles_posterior_weights1}) to (\ref{Eq: particles_posterior_weights3}) need to be changed to separately update survived and newborn particles. Details of the changed Equations can be found in \cite{ImprovedSMC2010}.

%a certain number of particles with a low weight are generated, each assigned with the ID of $\boldsymbol{z}_k$. The positions of the particles are sampled from a Gaussian distribution with a mean of $\boldsymbol{z}_k(1:3)$ and the covariance matrix $\boldsymbol{\Sigma}$ as described in Eq. (\ref{Eq: likelihood new}). The number of particles generated from $\boldsymbol{z}_k$ is determined by the intensity of the birth RFS $\gamma_{k|k-1}(\boldsymbol{x}_k)$ in Eq. (\ref{Eq: prediction SMC-PHD}). Further details regarding particle birth can be found in \cite{dspMap}. In contrast to \cite{dspMap}, the particles in this work incorporate an additional instance ID dimension. 
%The ID is assigned to be the same as the measurement point.

The resampling step is to mitigate the particle degeneracy problem and control the number of particles. We still use the rejection sampling \cite{RejectionSampling} approach for each voxel subspace in the map \cite{dspMap}. Particles in one voxel subspace are resampled by their weights, regardless of the instance ID. 
Particles with higher weights are more likely to survive or be duplicated in the resampling process, while the particles with lower weights are more likely to be removed, which is the ``death'' of the particles.
The overall weight of the particles in one voxel subspace is the same before and after resampling. Therefore, the resampling process does not affect the occupancy status estimation. \hl{Particles newly born in the same step are excluded from the resampling process.
To reduce computational cost, resampling is triggered to halve the number of particles only when a voxel becomes full, thereby freeing up space for the insertion of new particles.}

%If one voxel subspace contains particles with multiple IDs, the ID whose particles have the larger weights will be more likely to be assigned with more particles. 

%The map performs resampling only when a voxel subspace has a new particle to add for efficiency purpose. %Given that newborn particles originate from measurements, the resampling is carried out solely for the subspaces that contain the measurement points.

Considering Eq. (\ref{Eq: PHD_integral}) and (\ref{Eq: particles_posterior_weights3}-\ref{Eq: X sub-RFS}), the cardinality expectation of points that has $id = I$ in a voxel subspace $\mathbb{V}$ (constrained by position dimensions) is calculated by:

\begin{equation} \label{Eq: cardinality expectation of points in a voxel subspace}
  \mathbf{E}\left[ | \text{X}_I^{\mathbb{V}} |  \right] = \iiint_{\mathbb{V}} D_{\text{X}_I} \text{d} x \text{d} y \text{d} z = \sum_{i=1}^{L_{I}^{\mathbb{V}}} w^{(i)}_k 
\end{equation}
where the subscript or superscript $\mathbb{V}$ suggest the point or particle is in the voxel subspace. $L_{I}^{\mathbb{V}}$ is the number of particles with $id = I$ in the voxel subspace. Time step $k$ is omitted in the equation for simplicity. The cardinality expectation of all points in the voxel subspace is:
\begin{equation}\label{Eq: cardinality expectation of all points in a voxel subspace}
  \mathbf{E}\left[ | \text{X}^{\mathbb{V}} |  \right] = \sum_{I} \mathbf{E}\left[ | \text{X}_I^{\mathbb{V}} |  \right]
\end{equation}
%is calculated by summing the weights of particles within the voxel.
%The occupancy status is then determined by setting a threshold on the summation of the weights. If the summation exceeds the threshold, the voxel subspace is labeled as occupied. 
%If the summation exceeds a predefined threshold, the voxel subspace is labeled as occupied. 
We adhere to an ``occupancy first and ID second'' strategy, prioritizing the occupancy status over the object ID in navigation tasks. Therefore, the occupancy status of the voxel subspace is first estimated by applying a threshold on $\mathbf{E}\left[ | \text{X}^{\mathbb{V}} |  \right]$.
%\todo{Compare with the Pignistic probability in experiments and then modify this part.} 
If the voxel subspace is determined occupied, then its ID is estimated by finding the ID with the largest $\mathbf{E}\left[ | \text{X}_I^{\mathbb{V}} |  \right]$.

%finding the ID of the particles with the largest summation of weights.
%\todo{Add equations.}

\subsection{Memory Enhancement} \label{Section: Memory Enhancement}

Since objects with the same semantic label in one environment usually geometrically resemble each other, memory of previously observed objects can be used to conjecture the occluded portion of an object. 
For example, in Fig. \ref{Fig: memory} (a), the surface facing the camera is observed while the remaining area on the object is occluded. 
\hl{We store particles from previously observed objects as templates, which serve as a memory to account for the occluded parts of newly observed instances that have the same semantic label.}
%We aim to use previously observed information to conjecture the occluded portion.
This conjecture is important for navigation tasks because it can help the robot avoid planning in the space that is likely to be occupied. Additionally, it accelerates the map's response to previously observed objects by bearing particles to the occluded portion in advance. 
% In the following, we first introduce the structure to realize memory enhancement and then introduce the matching algorithm used in this process. 

\subsubsection{Structure} \label{Section: Memory Structure}
\begin{figure}[t]
  \centering
  \includegraphics[width=3.4in]{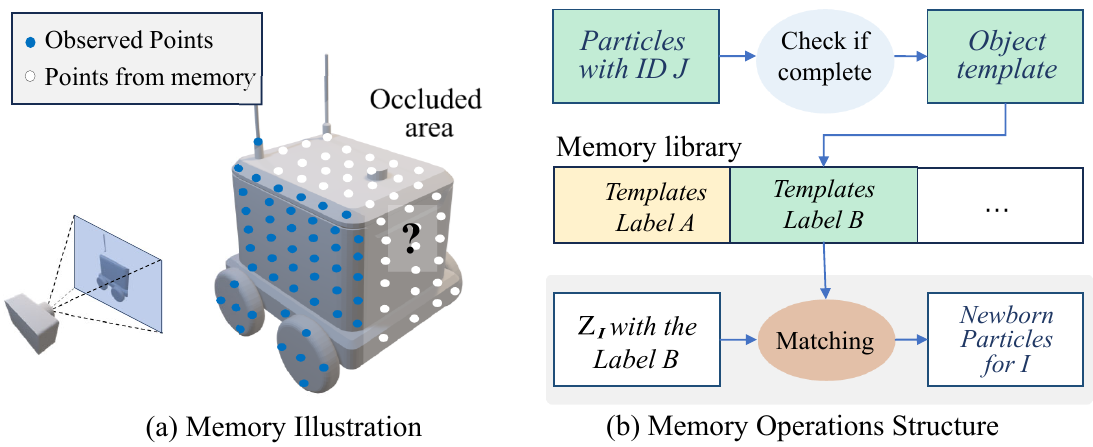}
  \caption{Illustration of the memory enhancement. In (a), the blue points are currently observed while the white points are occluded and are aimed to be conjectured. In (b), the memory enhancement structure is shown. The first row shows the process of adding a template in the memory library, which is shown in the middle row. The last row shows the process of matching the particles with a template in the memory library. The green background indicates that the particles or templates correspond to the same semantic label.}
  \label{Fig: memory}
\end{figure}

Fig. \ref{Fig: memory} (b) illustrates the structure of the memory enhancement module. The memory library is illustrated in the middle row of the figure. During the mapping process, we have particles with different instance IDs. Suppose an instance $J$ is well observed from various directions and is completely modeled. In that case, the particles with ID $J$ are stored as a template with the semantic label of $J$ in the memory library. Each label in the library can encompass several templates with distinct shapes.
In practical navigation scenarios, a robot rarely observes an object from all directions. 
We evaluate the completeness of the instance by uniformly sampling rays from the mass center of the voxels of this instance and calculating the percentage of rays that intersect with the voxels. If this percentage exceeds a predefined threshold, the instance is considered completely modeled, and its particles are stored as a template.

%The memory is integrated into the particle birth step. When a new instance with measurement points $\text{Z}_{I}$ is observed, we match the measurement points with the templates that have the same semantic label in the library. 
\hl{The memory is integrated into the particle birth step. When a new instance with measurement points $\text{Z}_{I}$ is observed and the number of measurement points exceeds a threshold (e.g., five thousand), we match these points with templates of the same semantic label and generate additional newborn particles based on the best-matched template.}
Let $\bar{\text{T}}_J = \{ P^{(1)}_J, P^{(2)}_J, \dots , P^{(L_T)}_J \}$ denote the template generated from instance $J$, containing $L_T$ particles. We then add $L_T$ newborn particles in addition to the newborn particles in Section \ref{Section: Birth and Resampling}.
Each newborn particle $P^{(i)}_b$ has the same position as the particle $P^{(i)}_J$ but weight $w_{b,k}$, which turns to
\begin{equation}
  w_{b,k} = \frac{v_{b,k|k-1}}{M_k L_b + L_T}
\end{equation}
%Note that these particles are conjectured particles without observation. They have the same positions as the particles in the template but a very low newborn particle weight. 
In the update step, the weight of these particles is updated with lateral observations, the same as the other particles, so that the conjecture can be corrected. If a voxel subspace is not determined occupied but contains non-updated conjectured particles, it is labeled ``speculatively occupied.''

% The memory is used in the particle birth step and the prediction step. In the particle birth step, when a new instance with measurement points $\text{Z}_{I}$ is observed, we match the measurement points with the templates that have the same semantic label in the library and generate newborn particles with the particles in the template, in addition to the newborn particles in Section \ref{Section: Birth and Resampling}. 
% In the prediction step, when an instance $J$ is re-tracked after occlusion, we match the previously updated particles with ID $J$ with the re-tracked measurement points $\text{Z}_J$ to move the particles to the correct position.

\subsubsection{Matching} \label{Section: Matching}
%PHD-based. Emphasize the importance of the efficiency. Matching can have errors. Discuss the robustness of the system.
The matching algorithm matches the particles in the template with the measurement points of a new instance. The matching algorithm should be efficient since there can be multiple new instances at a time. 
Unlike the matching between two regular point clouds, the matching between the particles and the measurement points should consider the nature of the particles representing the PHD and having different weights. 
In addition, the measurement points also contain hidden information, which is, the space between the camera and the measurement points should be free space. Therefore, we introduce a PHD-based matching algorithm to match the particles with the measurement points. 

The algorithm relies on a similarity score defined with the property of PHD described in Eq. (\ref{Eq: PHD_integral}). 
Suppose the measurement points in $\text{Z}_I$ are in a bounding box space $\mathbb{S}_I$, where $I$ is the instance ID, and the boundary of $\mathbb{S}_I$ can be found by searching the minimum and maximum coordinates of the measurement points. We divide $\mathbb{S}_I$ into voxel subspaces $\{\mathbb{S}_I^{(i)}, i \in N_I\}$, where $N_I$ is the number of subspaces. Suppose $h(\mathbb{S}_I^{(i)})$ is the expected point number in $\mathbb{S}_I^{(i)}$. 
If $\mathbb{S}_I^{(i)}$ contains at least one measurement point, $h(\mathbb{S}_I^{(i)})$ is considered as $1$. If $\mathbb{S}_I^{(i)}$ is observed to be free space (determined by raycasting), $h(\mathbb{S}_I^{(i)})$ is considered as $-1$. Otherwise, $h(\mathbb{S}_I^{(i)})=0$. 
Then we iterate over the voxel subspaces and calculate the similarity score with the $J_{th}$ template $\bar{\text{T}}_J$ using the following equation: 
\begin{equation} \label{Eq: similarity score}
  Score(\text{Z}_I, \bar{\text{T}}_J) = \frac{1}{N_I} \sum_{i=1}^{N_I} \min \{\int_{\mathbb{S}_I^{(i)}} D_{\bar{\text{T}}_J}(\boldsymbol{x}) \text{d}\boldsymbol{x}, 1\} \cdot h(\mathbb{S}_I^{(i)})
\end{equation}
where $D_{\bar{\text{T}}_J}(\boldsymbol{x})$ is the PHD of $\bar{\text{T}}_J$ at position $\boldsymbol{x}$. $\int_{\mathbb{S}_I^{(i)}} D_{\bar{\text{T}}_J}(\boldsymbol{x}) \text{d}\boldsymbol{x}$ is the PHD integral of $\bar{\text{T}}_J$ in $\mathbb{S}_I^{(i)}$ and equals the weight summation of the template particles in $\mathbb{S}_I^{(i)}$ according to Eq. (\ref{Eq: PHD_integral}) and (\ref{Eq: particles_posterior_weights3}). 
The integral and $h(\mathbb{S}_I^{(i)})$ are both normalized to one to avoid the influence caused by the voxel size, which is determined by a balance between the accuracy and the efficiency of the matching algorithm.
With the similarity score, the matching is performed using the RANSAC algorithm \cite{fischler1981random} to search for transformation matrices that maximize the score.

\section{Implementation Details} \label{Section: Implementation Details}
This section describes two implementation details of the proposed system: the data structure and the measurement points generation method. The former is to realize an efficient S$^2$MC-PHD filter, and the latter describes our choices of existing methods to implement the preprocessing modules.

\subsection{Data Structure} \label{Section: Data Structure}
\begin{figure}[t]
  \centering
  \includegraphics[width=3.4in]{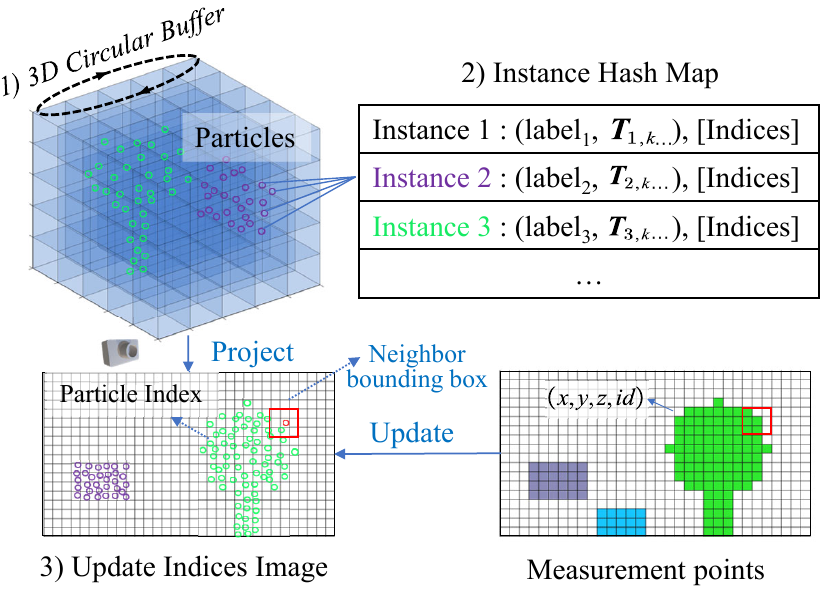}
  \caption{Data structure. Different colors of particles and measurement points represent different instance IDs. The blue object is newly observed. The red rectangle represents the neighbor bounding box area (activation space) of the pixel with a particle in red.}
  \label{Fig: data structure}
\end{figure}

As described in Section \ref{Section: World Model}, we use voxel subspaces to store the particles and use pyramid subspaces to distinguish the occluded area in the continuous space and accelerate the update process with an activation space as in \cite{dspMap}. To represent the voxel subspaces and the pyramid subspaces in practice, our previous work \cite{dspMap} uses a regular array and a dynamic vector, respectively, resulting in a high computational cost. To address this issue and consider the instance ID, we propose a new data structure composed of three parts: 1) a 3D circular buffer, 2) an instance hash map, and 3) an update indices image. The data structure is illustrated in Fig. \ref{Fig: data structure}.

\subsubsection{3D circular buffer} 
We store the particles in a 3D circular buffer indexed by Morton Code for efficiency. Each element in the buffer represents a voxel subspace and contains the particles within the subspace. 
With the circular buffer, only the indices of the voxel subspaces need to be updated when the robot moves. 
\hl{
Following Ewok\cite{Ringbuffer}, at the start of each time step, the indices are updated using the localization data, accounting for the relative motion of background particles to the sensor. For instances of interest, their particles are relocated to new voxels by recalculating their voxel indices based on their predicted positions, as described in Eq. (\ref{Eq: prediction object}).}
A position vector is maintained to compensate for position errors caused by the limited resolution of the voxel.

\hl{
Each voxel has a fixed particle capacity. If particles move to a voxel that has already reached its maximum capacity, resampling, as described in Section \ref{Section: Birth and Resampling}, is applied to this voxel to reduce the number of particles and free up space. If after resampling the space remains insufficient, the excess particles are discarded.}
%Each voxel has a fixed particle capacity. In case where particles move to a voxel that has already reached its maximum capacity, resampling, described in Section \ref{Section: Birth and Resampling}, is applied to reduce the number of particles and free up space.
%If, after resampling, the space remains insufficient, the excess particles are discarded.} 
Each particle contains the states described in Eq. (\ref{Eq: particle state vector}), namely position, weight, and instance ID, along with additional attributes such as a timestamp and a validity flag. The timestamp records the moment when the particle is updated with a measurement sharing the same ID, and it is used to calculate the forgetting function in Eq. (\ref{Eq: forgetting function}). The validity flag indicates whether the particle is valid, facilitating efficient particle deletion by setting the flag to false.

\subsubsection{Instance hash map}
A hash map is employed to store the instance ID along with the corresponding instance-level states. These states encompass each instance's semantic label, previous transformation matrices, and particle indices in the circular buffer. The particles' indices are used to quickly locate the particles in the buffer and predict the particles' positions in the prediction step (Section \ref{Section: Prediction}). The previous transformation matrices are used to predict a new transformation matrix in the prediction step when the instance is occluded. 

\subsubsection{Update indices image}\label{Section: Update indices image}
Since only a part of the particles in the circular buffer are visible to the camera and should be updated, we present an update indices image to leverage the Pinhole model to find these particles and store the indices. Specifically, a breadth-first search is used to identify the voxel subspaces in the FOV. In each voxel subspace in the FOV, the particles are projected to the image with the camera intrinsics and extrinsics. 
If the particle's depth is smaller than the depth of the corresponding measurement (indicating that the particle is not occluded), the particle's index is stored in a pixel.
In this context, each pixel serves a role similar to the pyramid subspace in \cite{dspMap}, as is illustrated by Fig. \ref{Fig: world model} (c). 
Subsequently, the weights of particles in each pixel are updated using only measurement points in a neighboring bounding box area, which act as the activation space in \cite{dspMap}, to accelerate the update process. In other words, the particles outside the bounding box area of a measurement point do not need to be updated with this measurement point. In Fig. \ref{Fig: data structure}, the red rectangle represents the neighbor bounding box area of the pixel with a particle in red. 
The size of the bounding box area is determined by the noise model and the distance from the camera to the measurement points. We present the derivation of the bounding box area in Appendix \ref{Appendix: Activation Bounding Box}.

%Particle birth and resampling steps are executed for the voxel subspaces. Whenever a particle is added or deleted, both the particle in the circular buffer and the instance hash map are updated.
% With the new data structure, the efficiency of the proposed map is improved although the additional instance ID dimension is added in the filter. We evaluate the computational time in Section \ref{Section: Experiment Real-time Performance}.

\subsection{\hl{Preprocessing}} \label{Section: Preprocess}
%Data association. Instance segmentation. Multi-object tracking. Transformation estimation.
Generating the measurement points requires preprocessing modules in Fig. \ref{Fig: system structure}. As discussed in Section \ref{Section: System Structure}, these preprocessing modules can be implemented using existing methods. %In our released framework, we use OpenVins \cite{geneva2020openvins} for localization. 
%The instance segmentation is realized with \cite{lyu2022rtmdet} and the dynamic objects are masked in OpenVins to better cope with the dynamic environment. PSPNet \cite{zhao2017pyramid}.
The panoptic segmentation is realized by Mask2Former \cite{cheng2022masked} using the OpenMMLab \cite{mmdetection} framework.
For object tracking and transformation estimation, we employ Superpoint \cite{detone2018superpoint} and Superglue \cite{sarlin2020superglue} to extract feature points for the instances and match them between two frames. Tracking is then implemented by voting the matched points, while transformation estimation is accomplished by applying RANSAC \cite{fischler1981random} on the matched feature points (considering depth) of each object. When an object is partially occluded, the transformation estimation can still be conducted with the matched feature points in the visible area. 
%In practice, when an object of interest is currently static, the transformation estimation result may still indicate slow movement due to depth noise. To address this issue, we employ a Bayesian filter to determine whether the object is moving. If the object is determined to be static at the moment, the filter sets its velocity to zero.

%We report the performance of the tracking and transformation estimation in Appendix \ref{Appendix: Tracking and Transformation Estimation Accuracy}.

% \todo{Add tracking and tracking result, moving or not determination, transformation matrix estimation method in appendix.}

\section{Experiment}  \label{Section: Experiment}
This section presents the experimental results. We first compare the proposed map with state-of-the-art mapping systems in terms of occupancy, semantic and instance estimation. The occupancy information is the foundation for the robot to realize safe navigation, while the semantics and instances of the occupied space are crucial for the robot to understand the environment. 
Then, the ablation study is conducted to compare the results of the proposed map with different update methods, and with and without memory enhancement.
Moreover, the efficiency of the proposed system is evaluated with computational time. Finally, real-world data is used to demonstrate the effectiveness of the proposed system in realistic scenarios.

% \todo{Change the expression of semantic noise to instance noise. Note by using external segmentation and tracking, the pose noise from inaccurate transformation estimation is also introduced.}

\subsection{Occupancy Estimation} \label{Section: Experiment Occupancy Estimation}
We adopt Virtual KITTI 2 \cite{gaidon2016virtual} \cite{cabon2020vkitti2} for evaluation. The cars, vans, etc., with possible dynamic motions in the dataset, are objects of interest. The dataset was chosen because it provides the ground truth about depth, panoptic segmentation, \hl{instance ID} and object poses over time, which is essential to creating the ground truth of occupancy and instance-aware semantics estimation in the dynamic environment. 
Creating the ground truth local instance-aware semantic occupancy map involves a two-step process.
The first step is accumulating the point cloud in the global coordinate using each frame's depth image, panoptic segmentation, and ego-pose, attributing semantic and instance IDs to individual points. As new frames emerge, points associated with the moving objects are transformed to their new positions using the ground truth object poses. 
Subsequently, at each time step $k$, the accumulated point cloud from $0$ to $k$ is divided into voxel subspaces and the occupancy status of each voxel in the local map range is determined by whether a point is present. 
The semantic label of the voxel subspace is determined by the majority of the points in the subspace.

%The method to create the ground truth is further introduced in \ref{Section: Experiment Occupancy Estimation}.

The occupancy estimation is evaluated using the Average Hausdorff Distance (AHD), the F1 score, and the Average Distance of movable objects (ADm). The AHD measures the average distance between the center points of the estimated occupancy voxels and those of the ground truth. A smaller AHD indicates better surface reconstruction performance. The F1 score is the harmonic mean of precision and recall. A higher F1 score shows better occupancy classification performance. 
The ADm is utilized to specifically measure the average Euclidean distance of the movable objects to the closest occupied voxels in the map. The movable objects in our scene are the aforementioned the objects of interests with possible dynamic motions. 
A smaller ADm indicates better mapping performance of these objects.
% The movable objects are possible dynamic objects of interests. 
% Due to different traffic conditions, the objects of interests may change from static to dynamic or reversely. 
% The Euclidean distance from the ground truth to the closest occupied voxels in the estimated map rather than the Hausdorff distance is calculated because some maps in comparison do not contain semantic information that segments the movable objects. 
Note the above metrics are evaluated for the local map with the ego-vehicle moving. Therefore, the metrics are calculated for each frame and then averaged over all frames. \hl{All the five sequences from Virtual KITTI 2 are used for evaluation and the results are averaged. These sequences contain $\{93, 17, 15, 21, 127\}$ vehicles, with the percentage of moving vehicles being $\{9.7, 52.9, 93.3, 100.0, 60.6\}$\%, respectively.}

The comparison is conducted with five state-of-the-art maps: ewok \cite{Ringbuffer}, k3dom \cite{DynamicMapICRA2021}, dsp map \cite{dspMap}, kimera-semantic \cite{rosinol2021kimera}, and voxblox++ \cite{grinvald2019volumetric}. Table \ref{Table: General Comparison} shows a general comparison of the maps.
In these maps, k3dom and dsp map are particle-based methods designed for dynamic environments, while Ewok uses raycasting for mapping without special consideration for dynamic objects, and none of the three methods considers semantics.
Kimera-semantic and voxblox++ are both TSDF-based semantic maps. Voxblox++ is instance-aware. In comparison, our map is an instance-aware semantic map and considers dynamic objects. Each map is evaluated in the situation with ground truth depth and noised depth based on a real-world noise model introduced in \cite{handa2014benchmark}. Furthermore, two cases are evaluated for the map with semantics: one with the ground truth semantics and tracking, and the other using the method in Section \ref{Section: Preprocess}. The lateral contains segmentation and tracking noise and is represented with a superscript $*$ in the result tables.

\begin{figure*}
  \centering
  \includegraphics[width=7.15in]{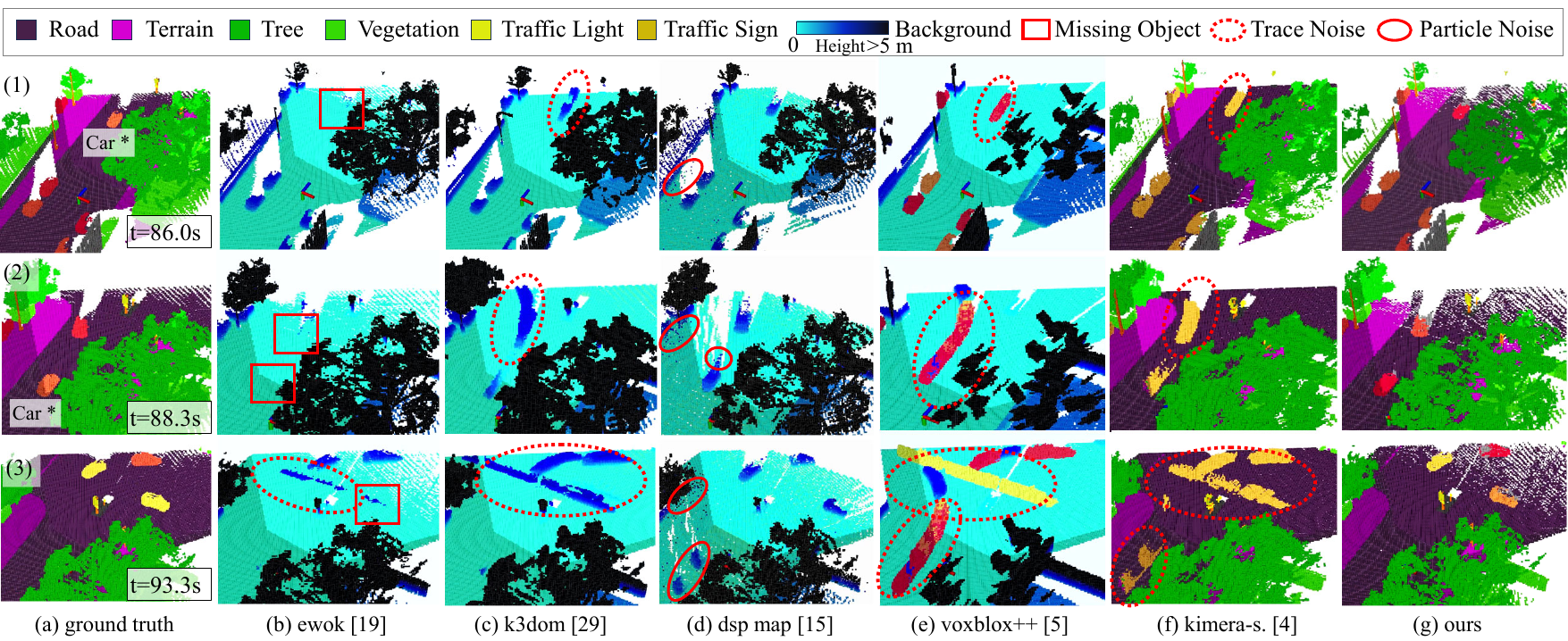}
  \caption{Mapping result comparison when using ground truth segmentation and tracking. The first column shows the ground truth map at three time steps, $t=\{ 86.0, 88.3, 93.3\}$s. The rest columns show the result of different maps, among which ours is presented in the last column. Voxels in the FOV are brighter than the rest to illustrate the currently observed area. 
  In (e) voxblox++ \cite{grinvald2019volumetric} and (g) ours, the vehicles are painted in random colors to show their instance-awareness.
  The meaning of the rest of the colors is illustrated in the legend above. If no semantic meaning is provided, the voxel is painted light blue to dark blue according to the height. Axes in the subfigures indicate the pose of the ego-vehicle. The red rectangles indicate the missing objects, while the red dashed ellipses illustrate the trace noise in the compared maps. The red solid ellipses in (d) suggest the noise in the occluded space caused by the individual particle motion model used in dsp map \cite{dspMap}.
  In Row (1) and (2) in Column (a), a car in orange is marked with ``Car$^*$''. The motion of this car causes the missing object or trace noise problem in the compared maps.}
  \label{Fig: result}
\end{figure*}

The voxel resolution in the test is 0.2 m, and the map size is (51.2, 51.2, 51.2) m (with $2^8$ voxels on each dimension). \hl{All the maps are compared in the voxelized form. For the TSDF-based maps\cite{rosinol2021kimera,grinvald2019volumetric}, a distance threshold is applied to determine the occupied voxels.}
\hl{We tested different distance thresholds with a sampling step of 0.05 m for the TSDF-based maps, and different occupancy thresholds with a sampling step of 0.1 for the rest maps to find the best performance of each map.} \hl{The remaining mapping parameters for our method are detailed in Appendix \ref{Appendix: Parameters}, while the parameters for other maps are kept at their default values as specified in their respective released code.}
%We tested different occupancy thresholds \hl{with a sampling step of 0.1 for maps that directly output voxel occupancy} \cite{Ringbuffer,DynamicMapICRA2021,dspMap}, and different distance thresholds \hl{with a sampling step of 0.05 m} for TSDF-based methods \cite{rosinol2021kimera,grinvald2019volumetric} to find the best performance of each map. 
\hl{The input images are from front camera with ID 0 in the dataset.}
The image size is $1242 \times 375$ pixels.
K3dom \cite{DynamicMapICRA2021} uses CUDA parallel computing and is tested with an NVIDIA RTX 2060S GPU. The rest of the maps are tested with an AMD Ryzen 9 5900X CPU with single-core computing.
Since kimera-semantic and voxblox++ are global maps, we crop out a local map to compare with the ground truth.
The results are shown in Table \ref{Table: Occupancy Estimation}. The arrows after each metric indicate the direction of the improvement. The best performance is in bold. Fig. \ref{Fig: result} presents an example mapping result of three time steps.

When using ground truth inputs, our map achieves the best performance in terms of all three metrics. The AHD and ADm are 73.8\% and 34.8\% smaller than the second-best map kimera-semantics, respectively, which indicates that our map has significantly improved the overall surface reconstruction and the movable object mapping performance in dynamic environments. 
The F1 score doesn't show a distinct difference between our map and the dsp map, but ours is still 4.1\% higher. From Fig. \ref{Fig: result}, it can be seen that our map does not suffer from the missing object problem and the trace noise problem as the other maps do. 

\begin{table}[htbp]
  \vspace{-1.0em}
  \caption{General comparison of the mapping systems.}
  \label{Table: General Comparison}
  \begin{center}
  \begin{tabular}{l|lccc}
  \hline
  \textbf{Map} & \textbf{Base} & \multicolumn{1}{l}{\textbf{Semantics}} & \multicolumn{1}{l}{\textbf{Instance}} & \multicolumn{1}{l}{\textbf{Dynamic}} \\ \hline
  ewok \cite{Ringbuffer}        & Raycasting           & No                                              & No                                          & No                                             \\
  k3dom \cite{DynamicMapICRA2021}       & Particles            & No                                              & No                                          & \textbf{Yes}                                   \\
  dsp map \cite{dspMap}     & Particles            & No                                              & No                                          & \textbf{Yes}                                   \\ 
  voxblox++ \cite{grinvald2019volumetric}   & TSDF                 & \textbf{Yes}                                    & \textbf{Yes}                                & No                                             \\
  kimera-s. \cite{rosinol2021kimera}      & TSDF                 & \textbf{Yes}                                    & No                                          & No                                             \\
  Ours         & Particles            & \textbf{Yes}                                    & \textbf{Yes}                                & \textbf{Yes}                                   \\ \hline
  \end{tabular}
\end{center}
\end{table}

\setlength{\tabcolsep}{5.5pt} % Reduce space for this table
\begin{table}[htbp]
  \vspace{-2.0em}
  \begin{threeparttable}
  \caption{Occupancy estimation comparison.}
  \label{Table: Occupancy Estimation}
  \begin{tabularx}{\columnwidth}{l|lll|lll}
  \hline
  \textbf{Depth Image} & \multicolumn{3}{c|}{GT Depth}           & \multicolumn{3}{c}{Depth With Noise}    \\ \hline
  \textbf{Map}         & \textbf{AHD}$\downarrow$   & \textbf{F1}$\uparrow$    & \textbf{ADm}$\downarrow$ & \textbf{AHD}$\downarrow$   & \textbf{F1}$\uparrow$    & \textbf{ADm}$\downarrow$ \\ \hline
  ewok \cite{Ringbuffer}                & 0.460          & 0.871          & 0.534          & 0.504          & 0.738          & 0.405          \\
  k3dom \cite{DynamicMapICRA2021}       & 1.006          & 0.643          & 0.425          & 3.858          & 0.467          & 0.508          \\
  dsp map \cite{dspMap}                 & 0.316          & 0.908          & 0.596          & 0.549          & \textbf{0.747} & 0.747          \\ 
  voxblox++ \cite{grinvald2019volumetric}           & 0.998         & 0.641          & 0.468          & 1.624          & 0.444          & 0.596          \\
  kimera-s. \cite{rosinol2021kimera}              & 0.256          & 0.878          & 0.423          & 0.461          & 0.634          & 0.364          \\
  \textbf{ours}                 & \textbf{0.067} & \textbf{0.945} & \textbf{0.276} & \textbf{0.195} & 0.689          & \textbf{0.242} \\ \hline
  voxblox++$^*$\cite{grinvald2019volumetric}          & 0.999          & 0.647          & 0.475          & 1.535          & 0.412          & 0.639          \\
  kimera-s.$^*$ \cite{rosinol2021kimera}            & 0.255          & 0.878          & 0.424          & 0.462          & 0.634          & 0.364          \\
  \textbf{ours}$^*$             & \textbf{0.066} & \textbf{0.945} & \textbf{0.318} & \textbf{0.201} & \textbf{0.682} & \textbf{0.278} \\ \hline
  \end{tabularx}
  \begin{tablenotes}
    \item $*$ indicates the case using segmentation and tracking from \ref{Section: Preprocess}.
  \end{tablenotes}
  \end{threeparttable}
\end{table}

\begin{figure}
  \centering
  \includegraphics[width=3.4in]{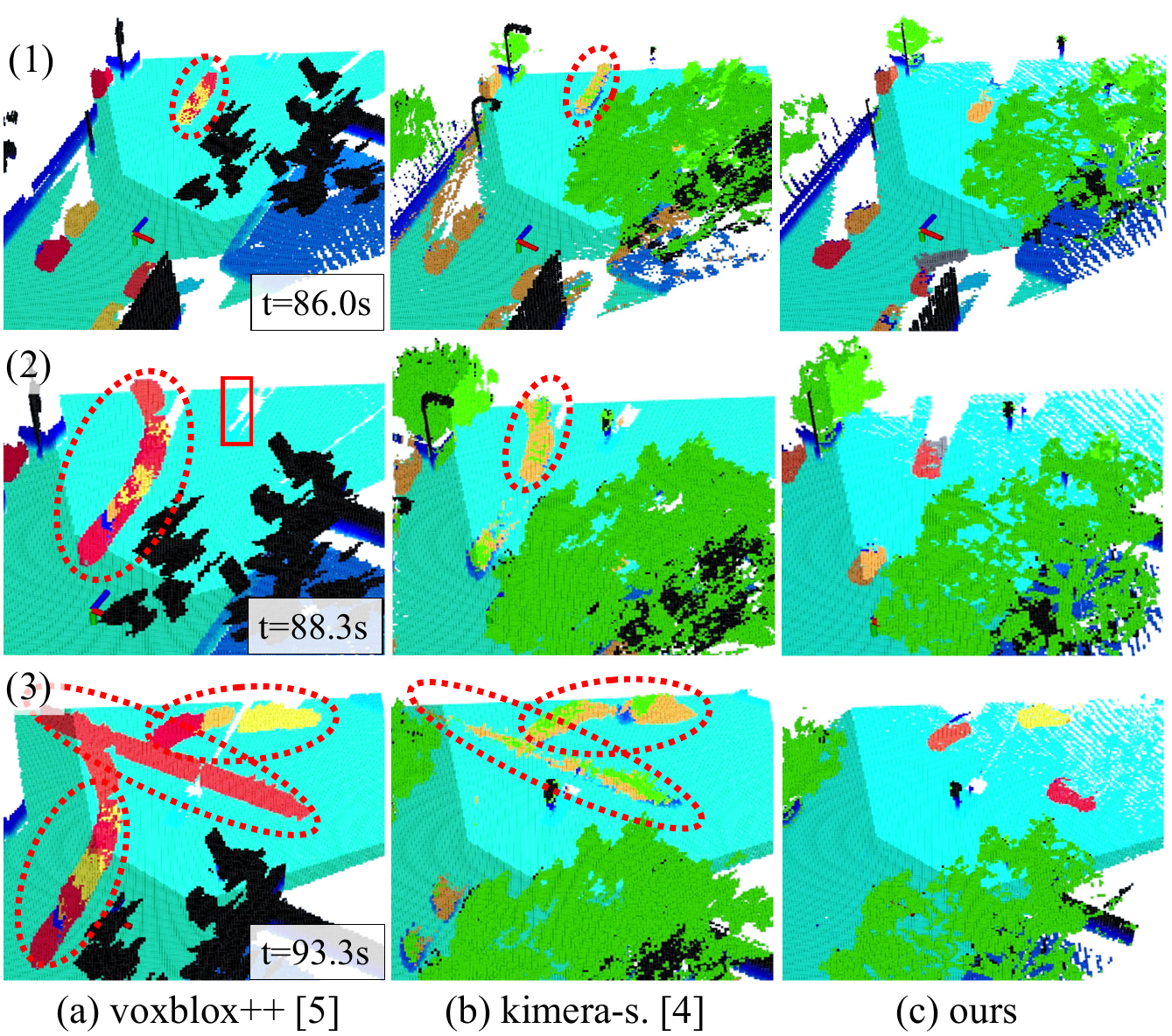}
  \caption{Mapping result comparison when segmentation and tracking are realized with the method described in Section \ref{Section: Preprocess}. Columns (a) to (c) show the mapping result of voxblox++ \cite{grinvald2019volumetric}, kimera-semantic \cite{rosinol2021kimera} and our map, respectively. The three time steps, the ground truth mapping result, and the legend are the same as in Fig. \ref{Fig: result}. voxblox++ \cite{grinvald2019volumetric} and kimera-semantic \cite{rosinol2021kimera} suffer from the trace noise problem. Voxblox++ \cite{grinvald2019volumetric} also misses details like the traffic sign and branches of the trees.}
  \label{Fig: external segmentation and tracking}
\end{figure}

When the depth with noise \cite{handa2014benchmark} is used, our map is the third best in terms of F1 score but is at least 57.7\% and 33.5\% better than other maps in terms of AHD and ADm, respectively. 
For the case that uses non-ground-truth segmentation and tracking (with superscript $*$), our map also shows a significant advantage over kimera-semantic and voxblox++. Fig. \ref{Fig: external segmentation and tracking} further illustrates the mapping result of the three maps using non-ground-truth segmentation and tracking.

%Compared with the case that uses ground truth segmentation and tracking, our map shows only a slight difference in terms of AHD and F1 score while the performance drop of ADm is also within 16\%, no matter the depth is with noise or not. In contrast, the performance of kimera-semantic drops significantly from 0.256 m to 1.264 m and 0.461 m to 1.308 in terms of AHD.

%This indicates that our map is robust to the noise brought by the utilized segmentation and tracking method.

Overall, our map shows the best performance in terms of the occupancy estimation and is more robust to the noise in depth image, and segmentation and tracking. 

\subsection{Semantic and Instance Estimation} \label{Section: Experiment Semantics Estimation}
Since our map is instance-aware, we evaluate both semantic segmentation and instance segmentation performance. The semantic segmentation performance is evaluated with 2D and 3D mean Intersection over Union (mIoU) of 15 classes in the Virtual KITTI 2 \cite{gaidon2016virtual} \cite{cabon2020vkitti2} dataset when ground truth segmentation is used. For the case that uses the OpenMMLab framework for segmentation, only the trees and cars are evaluated because the other classes are annotated differently or unavailable with the tested pre-trained segmentation model \cite{cordts2016cityscapes}.
In the 2D case, the labeled voxels in the map are projected to the image and compared with the ground truth segmentation image. In the 3D case, the labeled voxels are compared with the ground truth labeled voxels generated with the steps described in \ref{Section: Experiment Occupancy Estimation}. We present the results of static objects and movable objects separately in Table \ref{Table: Semantic Segmentation Static} and Table \ref{Table: Semantic Segmentation Movable} to show the performance of the map towards different types of objects. 
Voxblox++ \cite{grinvald2019volumetric} receives only the instance segmentations, which are available just for movable objects in the dataset, and thus, is not included in the Table \ref{Table: Semantic Segmentation Static}. In both tables, our map has the best performance regarding each metric. The advantage is distinctive regarding movable objects and the 3D mIoU metric. The 2D and 3D mIoU for movable objects are at least 45\% better than the second-best map, regardless of whether the depth, segmentation, and tracking have noise, in the tested cases.

% \begin{figure}
%   \centering
%   \includegraphics[width=3.1in]{figures/empty.pdf}
%   \caption{\todo{Add a radar plot.}  }
%   \label{Fig: radar plot}
% \end{figure}

\setlength{\tabcolsep}{7pt} % Reduce space for this table
\begin{table}[htbp]  
  \begin{center}
  \begin{threeparttable}
  \caption{Semantic segmentation results of static objects.}
  \label{Table: Semantic Segmentation Static}
  \begin{tabular}{l|cc|cc}
  \hline
  \textbf{Depth Image} & \multicolumn{2}{c|}{GT Depth} & \multicolumn{2}{c}{Depth With Noise} \\ \hline
  \textbf{Map}         & \textbf{mIoU}$\uparrow$     & \textbf{mIoU 3D}$\uparrow$   & \textbf{mIoU}$\uparrow$        & \textbf{mIoU 3D}$\uparrow$       \\ \hline
  kimera-s. \cite{rosinol2021kimera}            & 0.586             & 0.494              & 0.339                & 0.233                  \\
  \textbf{ours}        & \textbf{0.629}    & \textbf{0.834}     & \textbf{0.468}       & \textbf{0.322}         \\ \hline
  kimera-s.$^*$ \cite{rosinol2021kimera}            & 0.613             & 0.245              & 0.165                & 0.110                  \\
  \textbf{ours}$^*$        & \textbf{0.636}    & \textbf{0.473}     & \textbf{0.355}       & \textbf{0.245}         \\ \hline
  \end{tabular}
% \begin{tablenotes}
%   \item $*$ indicates the case using segmentation and tracking from \ref{Section: Preprocess}.
% \end{tablenotes}
\end{threeparttable}
\end{center}
\end{table}

\begin{table}[htbp]
  \vspace{-2.0em}
  \caption{Semantic segmentation results of movable objects.}
  \label{Table: Semantic Segmentation Movable}
  \begin{center}
  \begin{tabular}{l|cc|cc}
  \hline
  \textbf{Depth Image} & \multicolumn{2}{c|}{GT Depth}     & \multicolumn{2}{c}{Depth With Noise} \\ \hline
  \textbf{Map}         & \textbf{mIoU}$\uparrow$  & \textbf{mIoU 3D}$\uparrow$ & \textbf{mIoU}$\uparrow$    & \textbf{mIoU 3D}$\uparrow$  \\ \hline
  voxblox++ \cite{grinvald2019volumetric}           & 0.360          & 0.112            & 0.394            & 0.087             \\
  kimera-s. \cite{rosinol2021kimera}           & 0.464          & 0.212            & 0.365            & 0.101             \\
  \textbf{ours}        & \textbf{0.680} & \textbf{0.596}   & \textbf{0.660}   & \textbf{0.256}    \\ \hline
  voxblox++$^*$ \cite{grinvald2019volumetric}          & 0.364          & 0.112            & 0.382            & 0.085             \\
  kimera-s.$^*$ \cite{rosinol2021kimera}          & 0.462          & 0.161            & 0.353            & 0.085             \\
  \textbf{ours}$^*$       & \textbf{0.685} & \textbf{0.367}   & \textbf{0.591}   & \textbf{0.198}    \\ \hline
  \end{tabular}
\end{center}
\end{table}

The instance segmentation performance is evaluated with the mean F1 score of the instances in different frames. 
An instance in the ground truth and the estimated map is considered matched if the IoU is larger than 0.5. If no matching is found, the F1 score of the instance is 0. The results are shown in Table \ref{Table: Instance Segmentation}. Compared with voxblox++, our map shows over 60\% improvement in terms of 2D and 3D F1 score in all cases. 

Overall, our map has a significant advantage in terms of both semantic segmentation and instance segmentation.
%Comparing the results of different noise cases, 
The noise in depth image and segmentation and tracking affects more on the 3D metrics than the 2D metrics for all maps, which is reasonable because the noise in 2D is further amplified in 3D depending on the distance from the camera.

%indicates that building 3D semantic maps in real time with 2D measurements is still challenging.
%The performance when depth and segmentation and tracking are both with noise still has a large space for improvement. 

\begin{table}[htbp]  
  \caption{Instance segmentation results.}
  \label{Table: Instance Segmentation}
  \begin{center}
  \begin{tabular}{l|cc|cc}
  \hline
  \textbf{Depth Image} & \multicolumn{2}{c|}{GT Depth}         & \multicolumn{2}{c}{Depth With Noise}  \\ \hline
  \textbf{Map}                                 & \textbf{mF1}$\uparrow$ & \textbf{mF1 3D}$\uparrow$ & \textbf{mF1}$\uparrow$ & \textbf{mF1 3D}$\uparrow$ \\ \hline
  voxblox++ \cite{grinvald2019volumetric}                                   & 0.381            & 0.154              & 0.174            & 0.000              \\
  \textbf{ours}                                & \textbf{0.667}   & \textbf{0.588}     & \textbf{0.409}   & \textbf{0.186}     \\ \hline
  voxblox++$^*$ \cite{grinvald2019volumetric}                                   & 0.372            & 0.208              & 0.169            & 0.055              \\
  \textbf{ours}$^*$                                & \textbf{0.606}   & \textbf{0.402}     & \textbf{0.345}   & \textbf{0.177}     \\ \hline
  \end{tabular}
\end{center}
\end{table}

\subsection{Ablation Study} \label{Section: Experiment Ablation Study}
The presented results use the memory enhancement and the CF method in Section \ref{Section: Update}.
In this subsection, we further conduct an ablation study to compare the difference of using IF and CF methods for the update step and show the effect of using the memory enhancement. 
% The choice of the update method is to handle the noise in the instances, which are movable objects like cars in the dataset.

\begin{figure}
  \centering
  \includegraphics[width=3.5in]{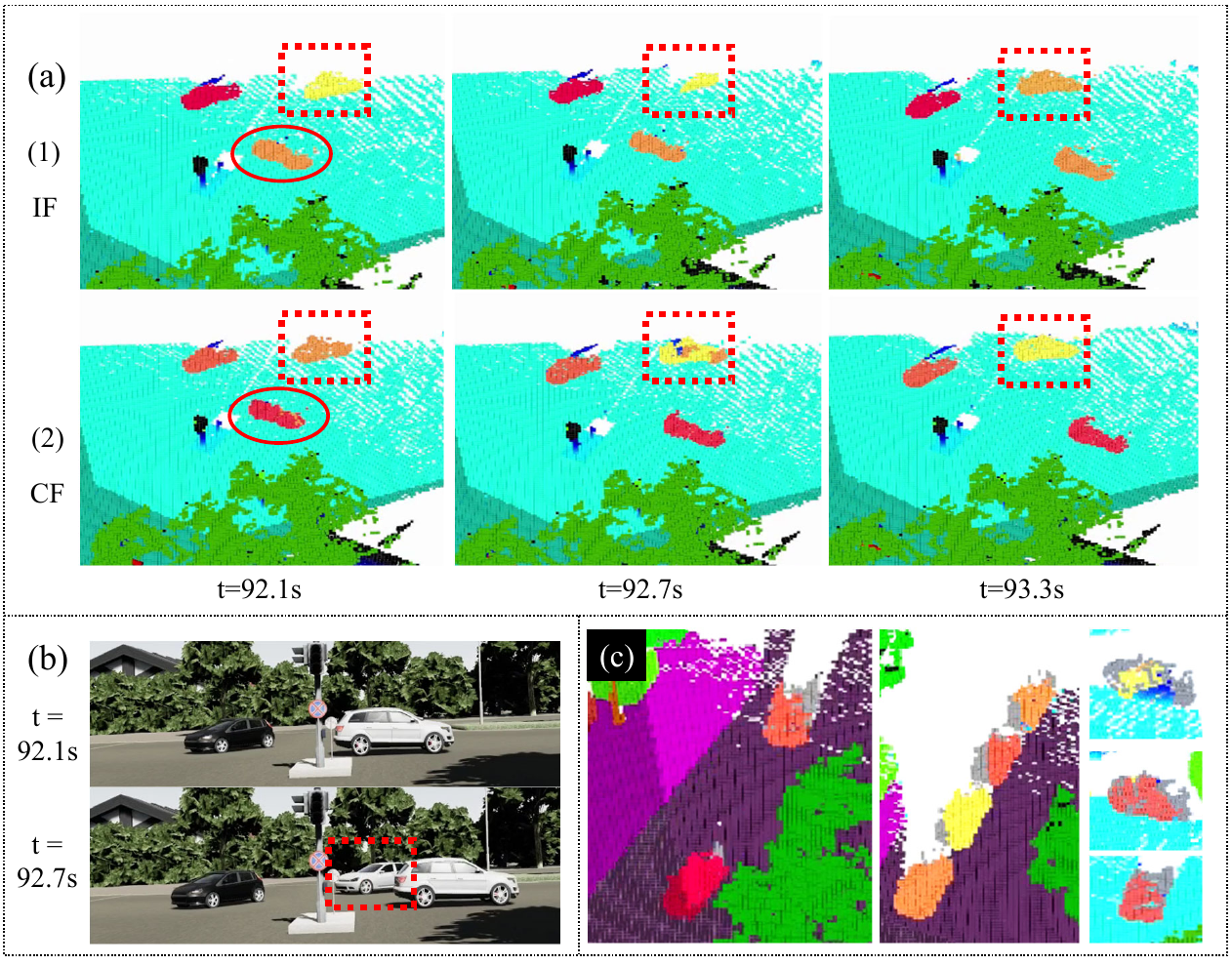}
  \caption{(a) Illustration of the missing object problem with the IF method. At 92.0 s, the car in the dashed red rectangle is occluded by the car in the red ellipse. At time 92.7 s, the former car is redetected and allocated with a new ID, which causes instance noise and is mostly missing in the map when using IF. In comparison, CF correctly updates the car. (b) shows the corresponding RGB images. The red rectangle indicates the occluded car. (c) presents several instances that have been matched with templates. The gray voxels show the voxels that are not observed but are ``speculatively occupied''.}
  \label{Fig: IF missing object problem}
\end{figure}

When the ground truth segmentation and tracking are used, the IF and CF methods show very similar performances. When the segmentation and tracking method in Section \ref{Section: Preprocess} is used, CF shows 7.1\% higher performance on ADm but 1.6\% lower performance on mF1 3D. The difference is insignificant because the instance noise takes a small portion of the instance estimations. 
However, if an instance in the measurement is allocated with a wrong ID due to missing detection, mislabeling or mismatching, the IF method suffers from the missing object problem that is highly detrimental to safety. An example mapping result is shown in Fig. \ref{Fig: IF missing object problem} (a). Therefore, CF is preferred in the proposed system.

\begin{figure}[h]
  \centering
  \includegraphics[width=3.4in]{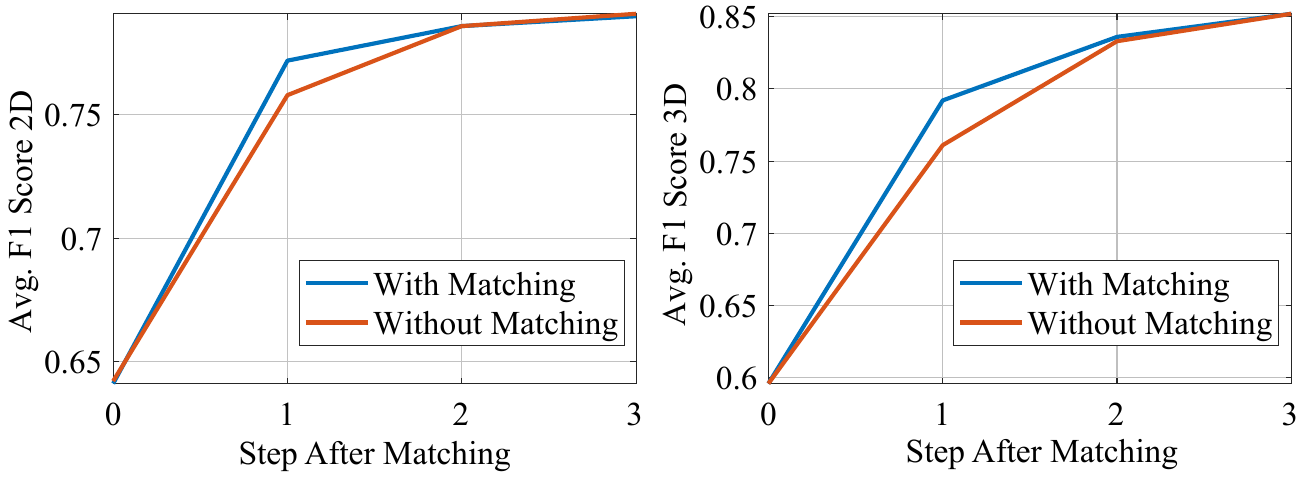}
  \caption{Response Comparison with and without Memory Enhancement.}
  \label{Fig: Memory Enhancement result}
\end{figure}

With memory enhancement, the map's response to the newly observed space on an instance is improved. Fig. \ref{Fig: Memory Enhancement result} shows the 2D and 3D mF1 score changing curve when the same instance is and is not matched with a template. At Step 0, a template is matched for the blue curve. Then, from Step 0 to 1, the F1 score increase of the blue curve is 12.9\% and 18.8\% faster than the red curve in 2D and 3D, respectively. The results indicate that memory enhancement is beneficial for the map in responding to the newly observed area on an instance. We illustrate some conjectured occupied voxels in Fig. \ref{Fig: IF missing object problem} (c) using the gray color. These conjectured voxels have also been proven helpful in motion planning \cite{wang2021learning,elhafsi_map-predictive_2020}. 

\subsection{Computation Time} \label{Section: Experiment Real-time Performance}
\begin{figure}[t]
  \centering
  \includegraphics[width=3.3in]{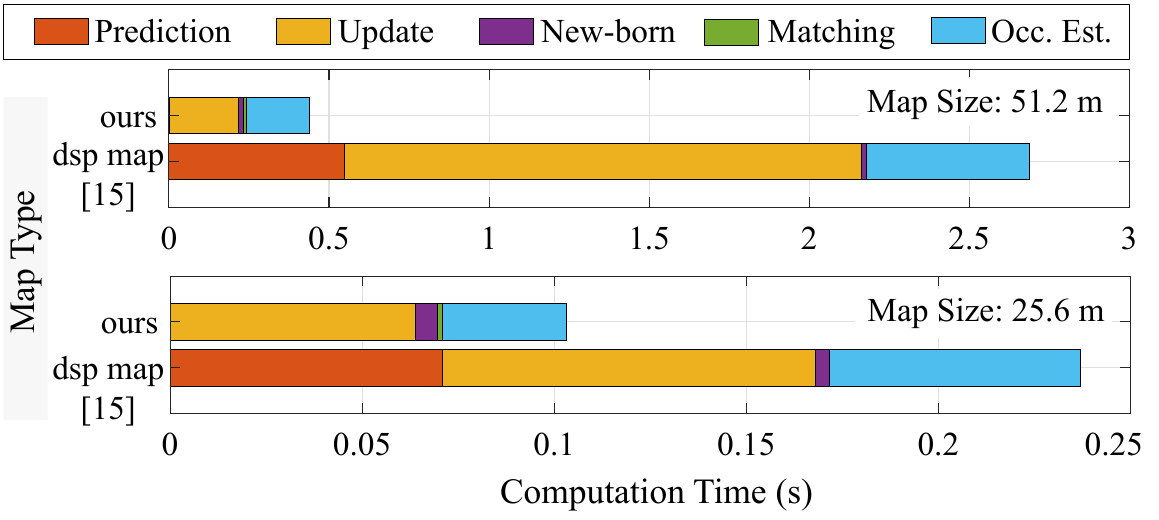}
  \caption{Average computation time two particle-based maps: our map and the dsp map \cite{dspMap}. When the map size is 25.6 m with voxel resolution 0.2 m, our mapping approach takes 103.2 ms per frame, reaching near real-time performance. The time consumption of different mapping steps is shown in different colors. The hardware is specified in \ref{Section: Experiment Occupancy Estimation}.}
  \label{Fig: run time}
\end{figure}

We compared the computational time of the proposed map and the dsp map \cite{dspMap}, in Fig. \ref{Fig: run time}. An additional map size of 25.6 m is tested, while the rest settings are the same as used in the previous tests. The hardware has been specified in \ref{Section: Experiment Occupancy Estimation}, and both maps use only a single core of the CPU. The time consumption of different mapping steps is shown in different colors. The preprocessing modules adopt existing works and can be adjusted with different models, and thus is not included.
It can be seen from Fig. \ref{Fig: run time} that the most time-consuming step is the update step. 
%The object update item in the figure represents the time used for multiple object tracking, transformation estimation, and velocity estimation in Fig. \ref{Fig: system structure} and is less than 1 ms. 
The template matching step in our map takes 9.8 ms with a big map size and 1.3 ms with a small map size.
In total, the proposed map takes 437.7 ms on average per frame when the map size is 51.2 m, which is six times faster than the dsp map. If the map size is reduced to 25.6 m, the proposed map takes 103.2 ms while the dsp map takes 236.9 ms. This demonstrates that the proposed map with the improved data structure described in Section \ref{Section: Data Structure}, our map is much more efficient than the dsp map though we have added more complex semantics and instance information.

% The raycasting-based map Ewok \cite{Ringbuffer} takes less than 100 ms per frame even with the big map size but is not designed for dynamic environments. Kimera-semantic \cite{rosinol2021kimera} and voxblox+ \cite{grinvald2019volumetric} are not compared since they are global maps. 
% K3dom \cite{DynamicMapICRA2021} is a particle-based dynamic local map, similar to dsp map and ours, while its time consumption is over 26 s per frame on RTX 2060S GPU. Therefore, our map is the most efficient dynamic map among the compared maps. It can reach real-time performance with the small map size but not with the big map size. For the application of mobile robots, a local map with size 25.6 m is sufficient while for the autonomous driving, the big map size is necessary. We will further improve the efficiency through parallel computing since particle-based methods have proven to be suitable for parallel computing \cite{RFSMap} \cite{DynamicMapICRA2021}.

Ewok \cite{Ringbuffer} takes less than 100 ms per frame to process, even with large map sizes, but it isn't suitable for dynamic environments and doesn't contain semantic information. 
Kimera-semantic \cite{rosinol2021kimera} and voxblox+ \cite{grinvald2019volumetric} are static global maps whose computation time increases from around 100 ms to over 1000 ms as the map grows. 
K3dom \cite{DynamicMapICRA2021}, a particle-based dynamic local map similar to the DSP map and ours, requires over 26 seconds per frame when conducting parallel computing on an RTX 2060S GPU when the map size is 25.6 m. Overall, our map emerges as the most efficient map among the compared dynamic maps and particle-based maps.

%It achieves real-time performance with the small map size, although not with the larger one. For mobile robot applications, a local map size of 25.6 meters is adequate, whereas autonomous driving requires larger map sizes. We plan to enhance efficiency through parallel computing, as particle-based methods are well-suited for this approach.
 
\subsection{Real-world Experiment} \label{Section: Experiment Real-world Experiment}
\begin{figure*}
  \centering
  \includegraphics[width=7.1in]{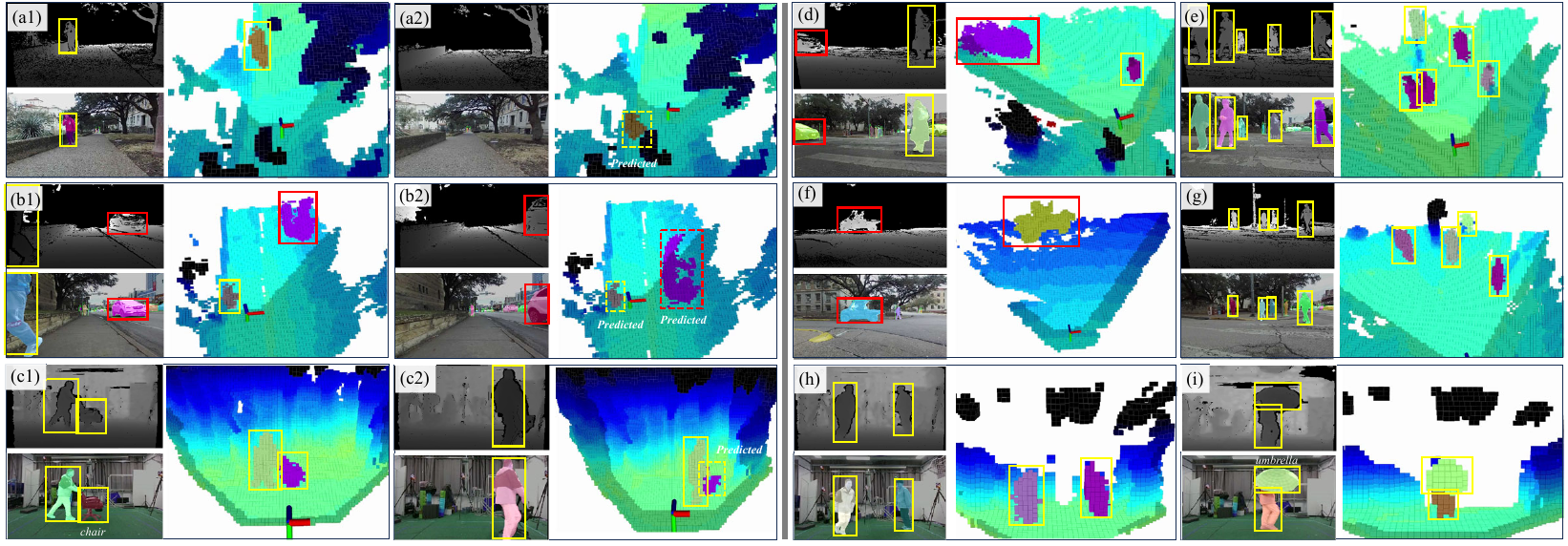}
  \caption{Mapping result in the real world. 
  On the left side of the figure, three scenes (a-c) are presented. Each scene has two frames, for example, (a1) and (a2), captured at two different time steps. In each frame, a depth image and an RGB image with segmentation masks are presented on the left, while the mapping result is shown on the right. 
  In the map view, the objects of interest are depicted in random colors, and the background is painted in green to light blue and then to dark blue according to the height. The camera's pose is illustrated by axes, with areas within the FOV appearing brighter than the surrounding areas. 
  In (a1), a human is running towards the camera and exits the FOV in (a2), where the map gives a predicted occupancy status of the human. Similarly, (b1) and (b2) show a scenario with a car and a human. In (c1), both the human and the chair are moving, while the chair is occluded in (c2), and the predicted occupancy status of the chair is shown. Red and yellow rectangles outline the cars and the other objects of interest, respectively, with dashed rectangles indicating the objects that are occluded or out of the FOV. Additional scenes are shown in (d-i) on the right side of the figure. Scenes (a-b) and (d-g) are from a ZED 2 camera used in the UT campus object dataset, while the rest are recorded by ourselves with a RealSense D455 camera.}
  \label{Fig: realworld test}
\end{figure*}

The real-world experiment is conducted with the RGB-D image pairs and pose estimation data from the UT campus object dataset \cite{zhang2023towards,coda2023tdr} and some additional RGB-D image pairs recorded by ourselves.
The image pairs in the UT campus object dataset came from a ZED 2 camera\footnote{ZED 2 Camera: \url{https://www.stereolabs.com/products/zed-2}} mounted on a mobile robot, while those recorded by ourselves came from a RealSense D455 camera.
Fig. \ref{Fig: realworld test} shows scenes of the original images and the constructed map, where the brighter part illustrate the area in the FOV. Scenes (a), (b), and (d-g) are from the UT campus object dataset and contain cars and people. Scene (c) and (h-i) are recorded by ourselves and contain people and objects moved by people, such as a chair and an umbrella. In scenes (a), (b) and (c), the first frame shows a time step when the moving objects can be observed, while the second frame shows a while later when the moving objects cannot be observed, our map gives a predicted occupancy status of the objects.
Although people have moving joints and are not rigid, we treat them as rigid bodies and consider the shape changes caused by the joint motions as noisy measurements.
Instance segmentation and tracking are realized with the method described in Section \ref{Section: Preprocess}. The computation frequency of the mapping component reaches 10 Hz with the hardware specified in \ref{Section: Experiment Real-time Performance} when the map size is 12.8 m.
More results, \hl{including real-time tests conducted in Delft using a ZED 2 camera}, can be found in the attached video. 
\hl{We also tested our method on the semantic mapping task of the KITTI-360 dataset\cite{liao2022kitti}. The results show that our method achieves the highest mIoU performance. Details can be found in Appendix \ref{Appendix: Additional Results}.
%In Appendix \ref{Appendix: Additional Results}, we also present the quantitative result for the semantic mapping task on the KITTI-360 dataset \cite{liao2022kitti}.
}
%The video of the tests is in \url{xxx}.

\section{Conclusion}  \label{Section: Conclusion}
This paper presents a dynamic instance-aware semantic map based on an S$^2$MC-PHD filter.
%Experimental results in the Virtual KITTI 2 dataset show that the proposed map has significant advantages over the state-of-the-art maps in terms of occupancy estimation, semantic segmentation, and instance segmentation. 
Experimental results on the Virtual KITTI 2 dataset demonstrate significant improvements in semantic and instance segmentation performance for possibly moving objects, such as cars, by over 45\% and 60\%, respectively, in both 2D and 3D, compared to SOTA maps. Moreover, the segmentation performance of static objects also excels, benefiting from the multi-hypotheses nature of the proposed method. Regarding occupancy estimation, our map outperforms SOTA maps by at least 30\% in terms of AHD and ADm under different noise conditions while maintaining a comparable F1 score. 
Additionally, an online memory enhancement module has been introduced and shown to improve the map's response to previously observed objects by 12.9\% and 18.8\% in 2D and 3D, respectively.
It is worth noting that the proposed method introduces a way to cohesively integrate the filtering-based mapping method with object motion prediction and memory-based geometric information conjecture, respectively, in the prediction and particle birth steps. 
While a constant velocity model and an online template matching method are used in the current implementation, extensions to learning-based motion prediction and 3D shape estimation or generation methods will be considered in future works, aiming to combine robustness and consistency inherent in the filtering-based method with the scalability and adaptability of learning-based methods.

% Moreover, the memory module provides an approach that combines the conjectured object geometric information with our filter-based map using the newborn particles. 
% Although our map can model the occupancy and semantics of dynamic objects, it uses a trivial constant velocity model to predict the future status of unobserved dynamic objects. 
% In future works, we will also consider the integrality of a more advanced context-aware motion model to improve prediction.

% Moreover, open-vocabulary semantics acquisition and update at the instance level will be considered to enhance the map's capability to handle more complex interaction-aware tasks.

{\appendices
\section{Activation Bounding Box} \label{Appendix: Activation Bounding Box}

The activation space is used to determine the range of pyramid subspaces whose particles should be updated with a measurement point \cite{dspMap}. The general idea is to ignore the particles in the pyramid subspaces that make the following condition true: for any particle $\tilde{\boldsymbol x}_{k}^{(i)}$ in the pyramid subspace, 
$g_{k}(\boldsymbol{z}_k|\tilde{\boldsymbol x}_{k}^{(i)}) \leq \epsilon$, where $g_{k}(\boldsymbol{z}_k|\tilde{\boldsymbol x}_{k}^{(i)})$ is the likelihood described in Eq. (\ref{Eq: likelihood new}) and $\epsilon \approx 0$ is a threshold.
As is mentioned in Section \ref{Section: World Model}, the Pinhole model is used to realize the pyramid subspace division. Thus, each pixel in the image plane corresponds to a pyramid subspace, as Fig. \ref{Fig: world model} (c) shows, and we can calculate an activation bounding box in the image plane to define the activation space. Any particle whose projection on the image plane is outside the bounding box can be ignored, while the weights of particles whose projections are inside the bounding box should be either increased or decreased in the update step.
We illustrate the Pinhole model and the activation bounding box in Fig. \ref{Fig: activation box}.

Since $F_{gt}(\tilde{\boldsymbol x}_{k}^{(i)}) \leq 1$ and $T_r(\boldsymbol{z}_k, \tilde{\boldsymbol x}_{k}^{(i)}) \leq 1$, the activation bounding box can be determined by finding the range of pixel position $(u, v)$, whose corresponding $\tilde{\boldsymbol x}_{k}^{(i)}$ can possibly satisfy $\mathcal{N}\left(\boldsymbol{z}_k; \tilde{\boldsymbol x}_{k}^{(i)}, \boldsymbol{\Sigma}\right) \leq \epsilon$, according to Eq. (\ref{Eq: likelihood new}).
Let $P = (x_1, y_1, z_1)$ and $Q = (x_2, y_2, z_2)$, $z_1, z_2 > 0$, be the position points of a measurement point $\boldsymbol{z}_k$ and a particle $\tilde{\boldsymbol x}_{k}^{(i)}$ in the camera frame, respectively. $P^\prime = (u_1, v_1)$ and $Q^\prime = (u_2, v_2)$ are the projections of $P$ and $Q$ on the image plane. 
As is used in \cite{dspMap}, we assume the covariance matrix $\boldsymbol{\Sigma}$ as a diagonal matrix with the same diagonal value $\rho(z_1)$, where $\rho(z_1)$ can be estimated by experiments with a sensor \cite{handa2014benchmark}. 
Then the particles that satisfy the condition $\mathcal{N}\left(\boldsymbol{z}_k; \tilde{\boldsymbol x}_{k}^{(i)}, \boldsymbol{\Sigma}\right) > \epsilon$ are in a sphere with radius $l$:
\begin{equation}
  l = \sqrt{2 \rho^2\left(z_1\right) \ln\left(\frac{1}{(2 \pi)^{\frac{3}{2}} \rho^3\left(z_1\right) \epsilon}\right)} 
\end{equation}

The projection of this sphere on the image plane can be proved to be an ellipse. The activation bounding box is calculated by finding the minimum and maximum values of $u_2$ and $v_2$ in this ellipse, given $P = (x_1, y_1, z_1)$ and $l$. Take $u_2$ as an example. Since $u_2 = f \frac{x_2}{z_2}$, where $f$ is the focal length, the minimum and maximum values of $u_2$ are equivalent to those of $\frac{x_2}{z_2}$. Considering the projection of the sphere on the x-z plane, $\frac{x_2}{z_2}$ can be represented by:
\begin{equation}
 %x_2 = x_1 + l sin(\alpha), \quad z_2 = z_1 + l cos(\alpha)\frac{x_2}{z_2}
 \text{f}(\alpha) = \frac{x_2}{z_2} = \frac{x_1 + l \sin(\alpha)}{z_1 + l \cos(\alpha)}
\end{equation}
where $\alpha$ is the angle shown in Fig. \ref{Fig: activation box}. By solving $\text{f}^\prime (\alpha) = 0$, the minimum and maximum values of $\text{f}(\alpha)$ is taken when 
\begin{equation}
 \alpha = 2 \arctan \left( \frac{x_1 \pm \sqrt{x_1^2 + z_1^2 - l^2} }{z_1-l} \right)
\end{equation}
if $x_1^2 + z_1^2 - l^2 \geq 0$ and $z_1 \neq l$. These conditions hold because otherwise, the origin is in the sphere, and $Q$ can be behind the camera. Then, the minimum and maximum values of $u_2$ can be calculated by $u_2 = f \cdot \text{f}(\alpha)$. The minimum and maximum values of $v_2$ can be calculated similarly.

\begin{figure}
  \centering 
  \includegraphics[width=3.0in]{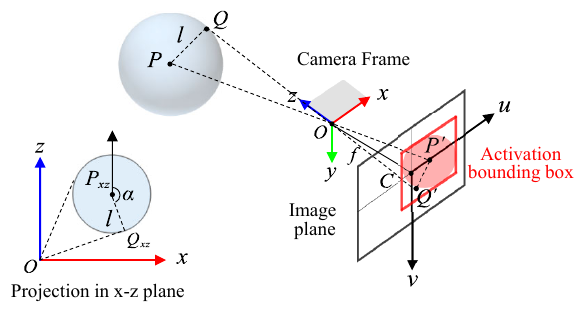}
  \caption{Illustration of calculating the activation bounding box when using the Pinhole model. $P$ is the position of a measurement point, and $Q$ represents the position of a particle. $P^\prime$ and $Q^\prime$ are the projections of $P$ and $Q$ on the image plane. $C$ is the center of the image. $f$ is the focal length of the camera. $l$ is the radius of the sphere which represents the surface where the Gaussian probability density is the same as a given threshold. 
  The activation bounding box of the measurement point is the bounding box of the sphere's projected ellipse on the image plane. A red rectangle is used to illustrate the bounding box. The projection of the sphere on the x-z plane is shown in the left bottom.}
  \label{Fig: activation box}
\end{figure}

\section{Additional Results} \label{Appendix: Additional Results}
\hl{
  This section presents the performance of our map in the semantic mapping task on the KITTI-360 dataset \cite{liao2022kitti}. The task evaluates global static semantic mapping performance across four test sequences using metrics such as accuracy (Acc.), completeness (Comp.), F1 score (F1), and mIoU over classes.
  To generate the global map, we assume all objects are static and accumulate the voxels from our map into a global map. Localization data is obtained from ORB-SLAM2 \cite{mur2017orb}, and the segmentation model used is CMNext \cite{zhang2023delivering}, with pretrained weights on KITTI-360. Depth images are generated using SGM \cite{hernandez2016embedded}, without applying any filters.
  The benchmark results are summarized in Table \ref{Table: kitti360}. Our map achieves the best performance in terms of mIoU, outperforming the second-best map by 5.11. Fig. \ref{Fig: kitti360} illustrates the result of our map on the Test Sequence 03.
}
  
\hl{
  We further conducted an ablation study by comparing the results of using our map with directly accumulating semantic points acquired using ORB-SLAM2, CMNext, and SGM, in the first global map of Sequence 0. Our map significantly improves accuracy from 29.6 to 72.8 and the F1 score from 44.9 to 76.5, demonstrating its denoising capability.
}

\begin{table}[h]
  \caption{\hl{Semantic Mapping Results in KITTI-360 Dataset\cite{liao2022kitti}}}
  \label{Table: kitti360}
  \begin{tabular}{l|l|l|l|l}
  \hline
  \textbf{Method}                    & \textbf{Acc.}       & \textbf{Comp.}   & \textbf{F1}             & \textbf{mIoU}    \\ \hline
  {\scriptsize ORB-SLAM2\cite{mur2017orb} + PSPNet\cite{zhao2017pyramid}}    & 81.77          & \textbf{74.89} & \textbf{78.15} & 32.48          \\
  {\scriptsize SUMA++ \cite{8967704} }                 & \textbf{90.98} & 64.19          & 75.27          & 19.40          \\
  {\scriptsize Ours + Preprocessing \cite{mur2017orb,zhang2023delivering} } & 79.15          & 72.45          & 75.64          & \textbf{37.59} \\ \hline
  \end{tabular}
\end{table}

\begin{figure}
  \centering 
  \includegraphics[width=3.5in]{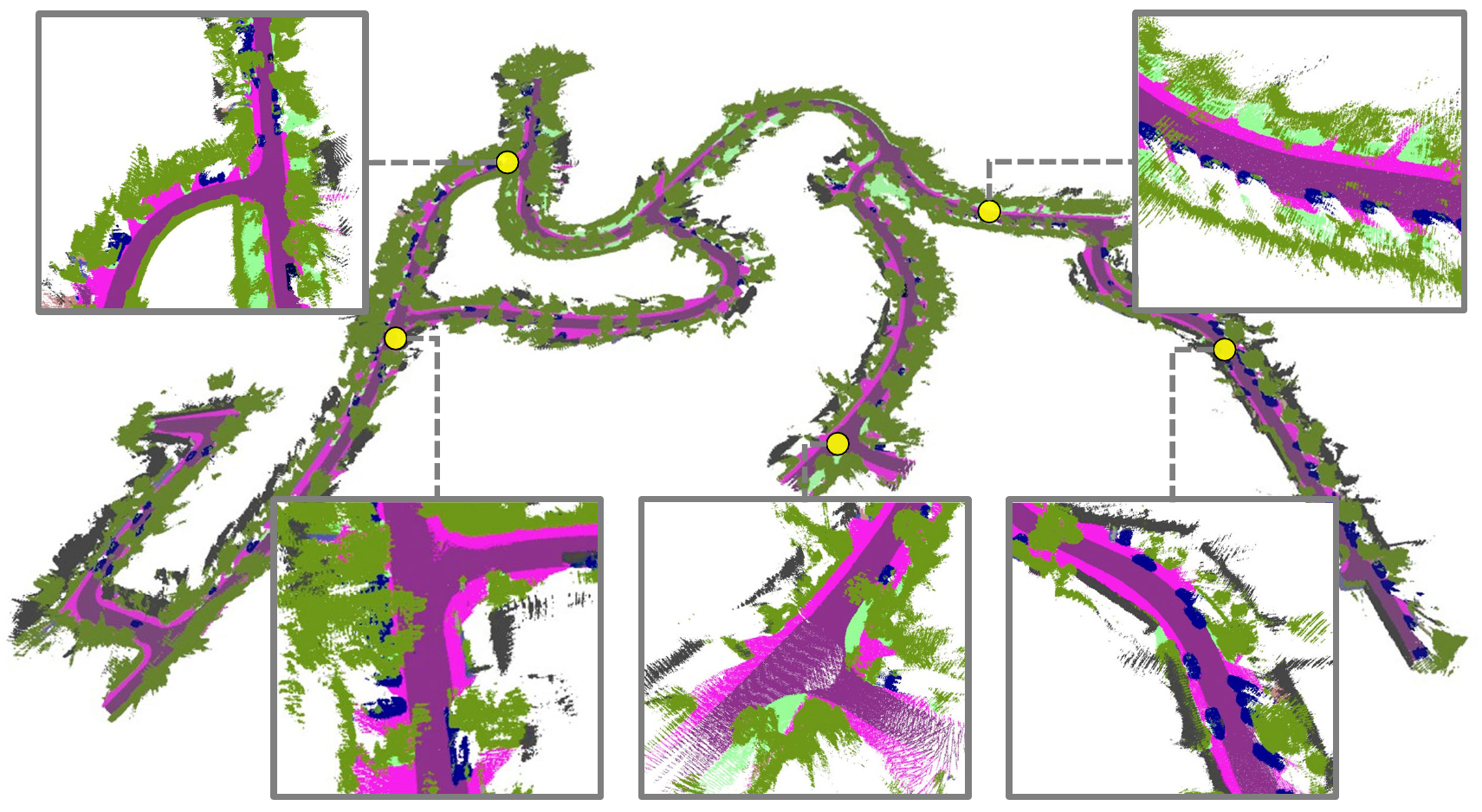}
  \caption{\hl{The semantic mapping result of our method on Test Sequence 03 of the KITTI-360\cite{liao2022kitti} semantic mapping challenge. 
  Five specific locations, marked with yellow spots, are enlarged and illustrated for a detailed view.
  The color of each label follows the scheme used in the Cityscapes dataset\cite{cordts2016cityscapes}.}}
  \label{Fig: kitti360}
\end{figure}

\section{Parameters} \label{Appendix: Parameters}
\hl{
The parameters used in the experiments are listed in Table \ref{Tabel: Parameters}. To enhance computational efficiency, several simplifications were made: the clutter intensity $\kappa_k(\boldsymbol{z}_k)$ was simplified to a constant $\kappa_k$, the prediction covariance $\boldsymbol{Q}$ was simplified to a constant diagonal matrix, and the positional noise covariance $\boldsymbol{\Sigma}$ of the depth camera was simplified to a diagonal matrix that changes linearly with the depth $d$. 
${Thr}_{score}$ represents the matching score threshold used in Section \ref{Section: Matching}. $N_{pix}^{bbox}$ denotes the pixel number from the center of the neighbor bounding box in Section \ref{Section: Update indices image} to its edges. $l_{voxel}$ is the size of the voxel subspace, and $N_{max}^{\mathbb{V}}$ specifies the maximum number of particles per voxel.

The occupancy threshold ${Thr}_{score}$ was determined through uniform sampling, as described in Section \ref{Section: Experiment Occupancy Estimation}, and was identified to be 0.2 for the tests with depth noise. 
The rest of the parameters were tuned based on experience. Specifically, $\kappa_k$, $\boldsymbol{\Sigma}$, and $\boldsymbol{Q}$ need to be adjusted to account for different depth camera noise characteristics. 
In the tests with depth noise, $\kappa_k = 0.4$, $\boldsymbol{\Sigma} = (0.03 d +  0.1) \boldsymbol{I}$, and $\boldsymbol{Q} = 0.05 \boldsymbol{I}$ were used. In real-world experiments (Section \ref{Section: Experiment Real-world Experiment}), $\kappa_k$ and $\boldsymbol{\Sigma}$ were further adjusted to $0.5$ and $(0.02 d + 0.3) \boldsymbol{I}$, respectively. 
Since the panoptic segmentation and tracking errors usually remain consistent between simulation and real-world environments, parameters related to these processes, such as $P_{tr}$, $S$, and $\Delta \bar{k}$, were not adjusted.
}

\begin{table}[t]
\caption{\hl{Parameters used in experiments}}
\label{Tabel: Parameters}
\resizebox{\linewidth }{!}{
\begin{tabular}{ll|ll|ll}
\hline
\textbf{Param.} & \textbf{Value} & \textbf{Param.}  & \textbf{Value}               & \textbf{Param.} & \textbf{Value} \\ \hline
$P_d$        & 0.98  & $P_s$         & 1                   & $\kappa_k$     & 0.01  \\ 
$\boldsymbol{Q}$         & 0.01$\boldsymbol{I}$ & $\boldsymbol{\Sigma}$      & ($10^{-3}d+10^{-2}$)$\boldsymbol{I}$ & $P_{tr}$       & 0.5   \\ 
$S$         & 1     & $\Delta \bar{k}$   & 5                   & $L_b$        & 5     \\ 
$w_{b,k}$       & 0.001 & ${Thr}_{score}$ & 0.6                 & $l_{voxel}$    & 0.2 m  \\ 
${Thr}_{occ}$  & 0.8   & $N_{max}^{\mathbb{V}}$     & 8                   & $N_{pix}^{bbox}$    & 5 pix   \\ \hline
\end{tabular}}
\end{table}

\bibliographystyle{IEEEtran}

\bibliography{head}

\begin{IEEEbiography}[{\includegraphics[width=1in,height=1.25in,clip,keepaspectratio]{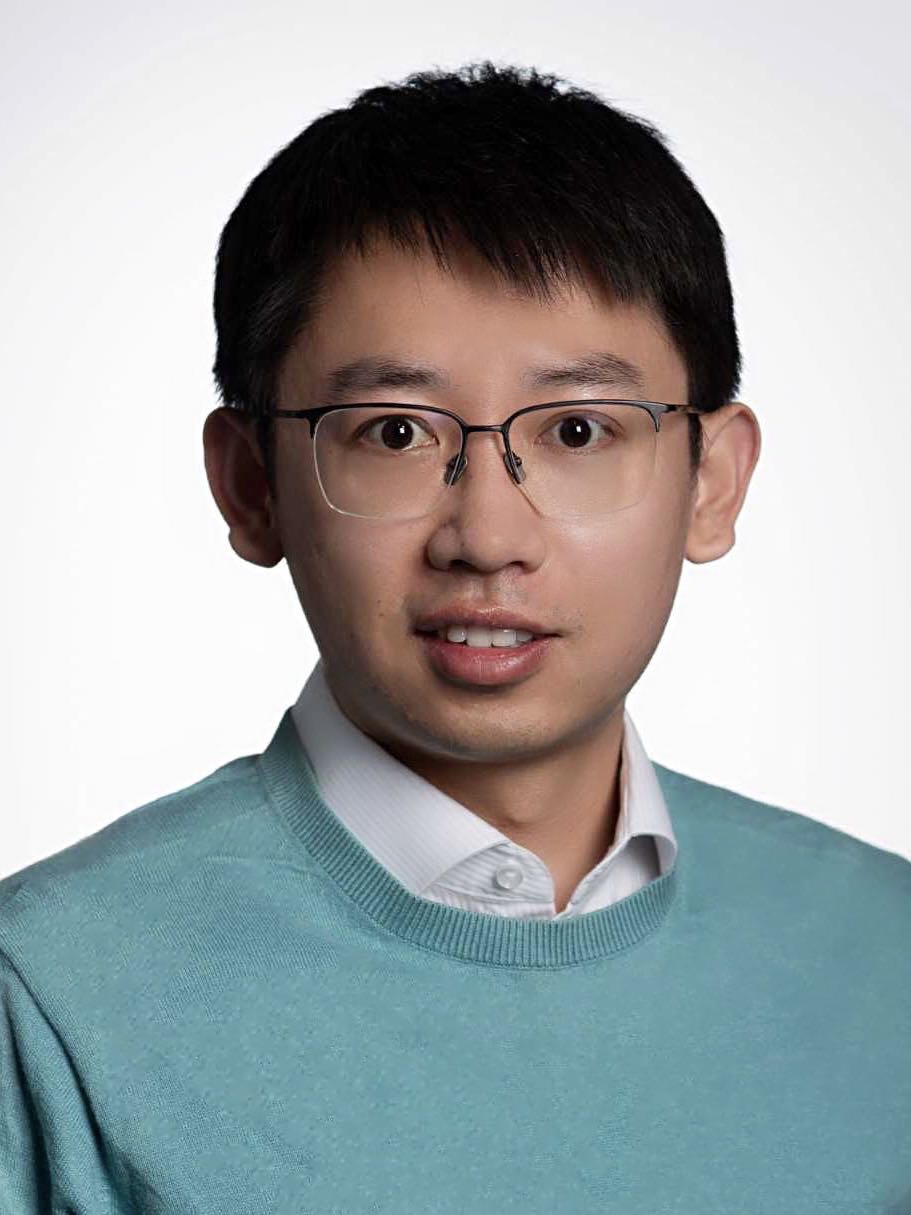}}]{Gang Chen}
received the B.E. degree and Ph.D. degree in mechanical engineering from Shanghai Jiao Tong University, Shanghai, China, in 2016 and 2022, respectively. He is currently a Postdoc researcher at the Cognitive Robotics Department, Delft University of Technology, the Netherlands. His research interest is in perception and perception-aware planning for the navigation of single- and multi-robot systems, with a special focus on dynamic environments.
\end{IEEEbiography}

\begin{IEEEbiography}[{\includegraphics[width=1in,height=1.25in,clip,keepaspectratio]{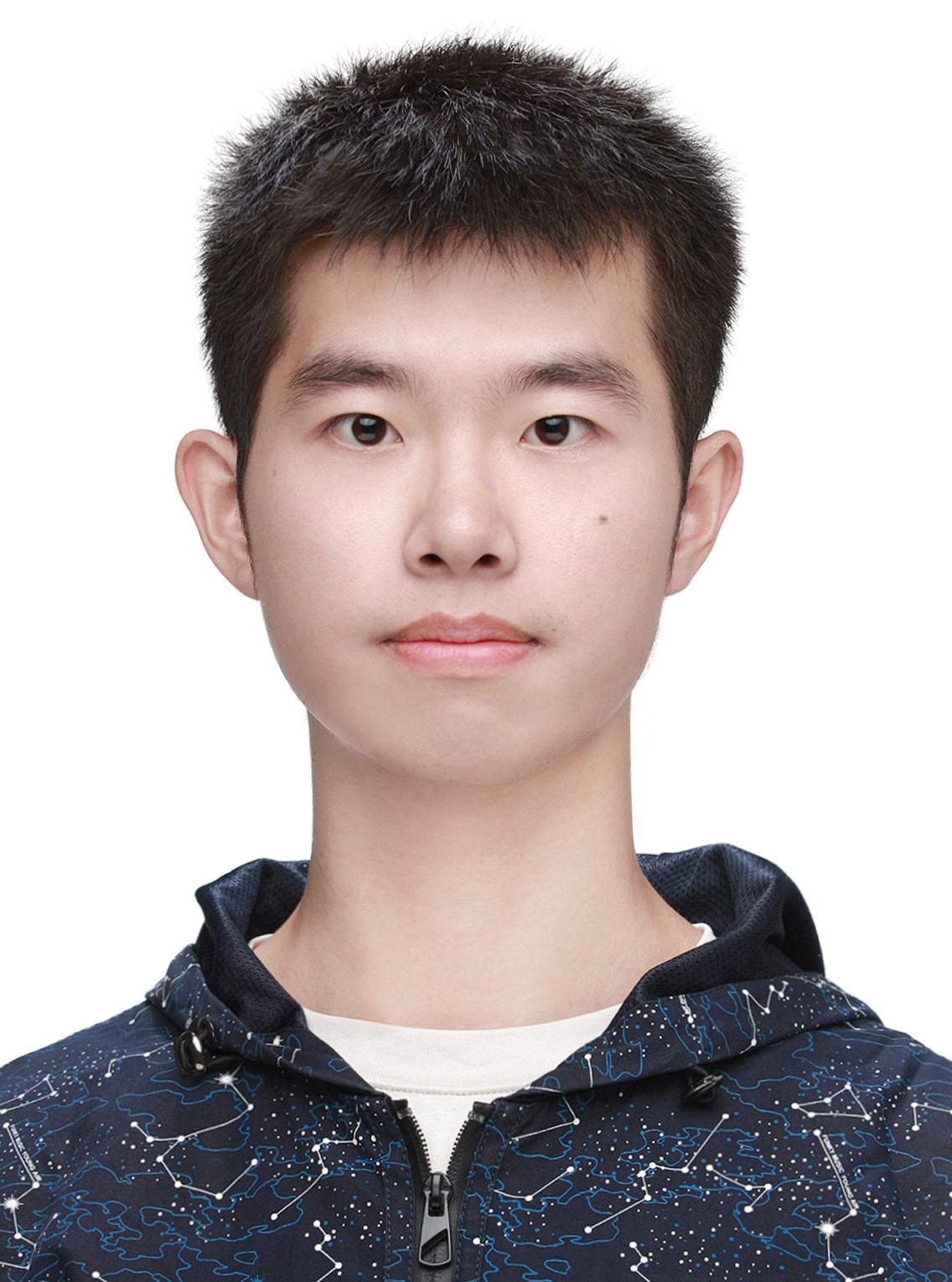}}]{Zhaoying Wang} received the B.E. degree in mechanical engineering from Wuhan University, Wuhan, China, in 2019. He is now a Ph.D. candidate at the State Key Laboratory of Mechanical System and Vibration, Shanghai Jiao Tong University. His research interests are Visual Inertial Odometry, SLAM, and multi-robot relative localization and mapping.
\end{IEEEbiography}

\begin{IEEEbiography}[{\includegraphics[width=1in,height=1.25in,clip,keepaspectratio]{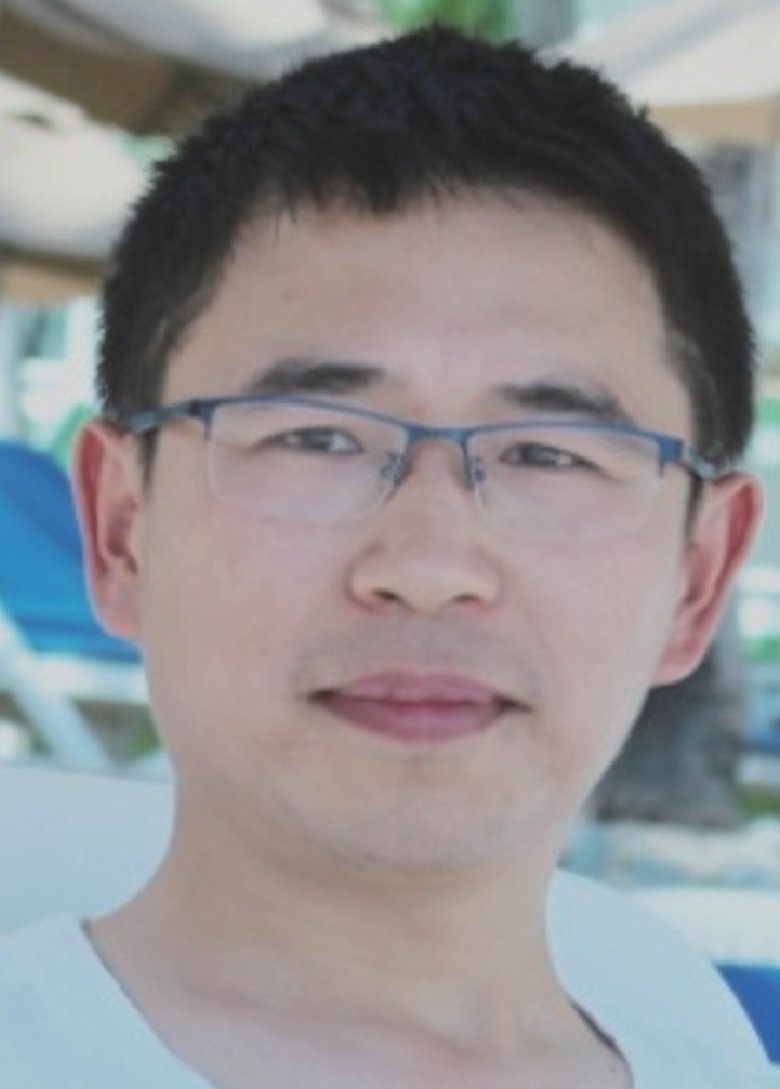}}]{Wei Dong} received the B.S. degree and Ph.D. degree in mechanical engineering from Shanghai Jiao Tong University, Shanghai, China, in 2009 and 2015, respectively. He is currently a tenured associate professor in the Robotic Institute, School of Mechanical Engineering, Shanghai Jiao Tong University. In 2022, he was selected into the Shanghai Rising-Star Program for distinguished young scientists. His research interests include active perception and cooperation of unmanned systems.
\end{IEEEbiography}

\begin{IEEEbiography}[{\includegraphics[width=1.1in,height=1.5in,clip,keepaspectratio]{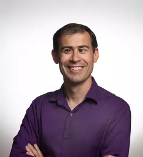}}]{Dr. Javier Alonso-Mora} is an associate professor at the Cognitive Robotics department of the Delft University of Technology, where he leads the Autonomous Multi-Robots Lab. Before joining TU Delft, Dr. Alonso-Mora was a postdoctoral associate at the Massachusetts Institute of Technology (MIT). He received his Ph.D. degree in robotics from ETH Zurich.
 
His main research interest is in navigation, motion planning, learning, and control of autonomous mobile robots, and teams thereof, that interact with other robots and humans in dynamic and uncertain environments. He is the recipient of multiple awards, including the IEEE Transactions on Automation Science and Engineering Best Paper Award (2024), an ERC Starting Grant (2021), and the ICRA Best Paper Award on Multi-Robot Systems (2019). He serves as an Associate Editor for the IEEE Transactions on Robotics and for Springer Autonomous Robots.
\end{IEEEbiography}

\end{document}